%% file: main.tex
\documentclass{article}

\usepackage[final]{neurips_2025}

\usepackage[utf8]{inputenc} % allow utf-8 input
\usepackage[T1]{fontenc}    % use 8-bit T1 fonts
\usepackage{hyperref}       % hyperlinks
\usepackage{url}            % simple URL typesetting
\usepackage{booktabs}       % professional-quality tables
\usepackage{amsfonts}       % blackboard math symbols
\usepackage{nicefrac}       % compact symbols for 1/2, etc.
\usepackage{microtype}      % microtypography
\usepackage{xcolor}         % colors

\usepackage{amsmath}
\usepackage{amssymb}
\usepackage{mathtools}
\usepackage{amsthm}

\usepackage{graphicx}  
\usepackage{subcaption}  
\usepackage{caption}  
\usepackage{adjustbox}
\usepackage{makecell}

\usepackage{colortbl}
\definecolor{highlight}{rgb}{0.92,0.97,1}
\usepackage[all]{nowidow}  
\usepackage{authblk}
\usepackage{makecell}
\title{Homogeneous Keys, Heterogeneous Values: Exploiting Local KV Cache Asymmetry for Long-Context LLMs}

\author[*,+]{\textbf{Wanyun~Cui}}
\author[*]{\textbf{Mingwei~Xu}}

\affil[*]{Shanghai University of Finance and Economics}
\affil[+]{MoE Key Laboratory of Interdisciplinary Research of Computation and Economics\\ Shanghai University of Finance and Economics}
\affil[*]{cui.wanyun@sufe.edu.cn, mingweixu@stu.sufe.edu.cn}

\begin{document}

\maketitle

\begin{abstract}
Recent advances in Large Language Models (LLMs) have highlighted the critical importance of extending context length, yet the quadratic complexity of attention mechanisms poses significant challenges for efficient long-context modeling. KV cache compression has emerged as a key approach to address this challenge. Through extensive empirical analysis, we reveal a fundamental yet previously overlooked asymmetry in KV caches: while adjacent keys receive similar attention weights ({\it local homogeneity}), adjacent values demonstrate distinct {\it heterogeneous} distributions. This key-value asymmetry reveals a critical limitation in existing compression methods that treat keys and values uniformly. To address the limitation, we propose a training-free compression framework (AsymKV) that combines homogeneity-based key merging with a mathematically proven lossless value compression. Extensive experiments demonstrate that AsymKV consistently outperforms existing long-context methods across various tasks and base models. For example, on LLaMA3.1-8B, AsymKV achieves an average score of 43.95 on LongBench, surpassing SOTA methods like H$_2$O (38.89) by a large margin.Our code can be found in this \href{https://github.com/the-scale-lab/Asymkv}{link}.
\end{abstract}

\input{intro}

\input{related}

\input{method}

\input{exp}
\input{conclu}

\bibliographystyle{plain}
\bibliography{homogeneous}

\newpage
\section*{NeurIPS Paper Checklist}

\begin{enumerate}

    \item {\bf Claims}
        \item[] Question: Do the main claims made in the abstract and introduction accurately reflect the paper's contributions and scope?
        \item[] Answer: \answerYes{} % Replace by \answerYes{}, \answerNo{}, or \answerNA{}.
        \item[] Justification: 
        The abstract and introduction clearly state:
        \begin{itemize}
            \item Discovery of KV cache asymmetry (key homogeneity vs. value heterogeneity).
            \item Proposal of AsymKV, a training-free compression framework combining key merging and lossless value representation.
            \item Experimental validation showing SOTA performance (e.g., 43.95 vs. 38.89 for H$_2$O on LongBench).
        \end{itemize}

    \item {\bf Limitations}
        \item[] Question: Does the paper discuss the limitations of the work performed by the authors?
        \item[] Answer: \answerYes{} % Replace by \answerYes{}, \answerNo{}, or \answerNA{}.
        \item[] Justification: Please see \S~\ref{sec:conclu}. We discussed the potential engineering adaption problem.
    
    \item {\bf Theory assumptions and proofs}
        \item[] Question: For each theoretical result, does the paper provide the full set of assumptions and a complete (and correct) proof?
        \item[] Answer: \answerYes{} % Replace by \answerYes{}, \answerNo{}, or \answerNA{}.
        \item[] Justification: The paper includes detailed mathematical derivations for its key compression (Eq. 5) and lossless value compression (Eq. 8) methods. Assumptions, such as the use of the Fisher information matrix as an approximation for the Hessian, are explicitly stated. Complete proofs are provided in the main text and appendices, supported by references to established techniques like Taylor expansion.
    
    \item {\bf Experimental result reproducibility}
        \item[] Question: Does the paper fully disclose all the information needed to reproduce the main experimental results of the paper to the extent that it affects the main claims and/or conclusions of the paper (regardless of whether the code and data are provided or not)?
        \item[] Answer: \answerYes{} % Replace by \answerYes{}, \answerNo{}, or \answerNA{}.
        \item[] Justification: The experimental setup is thoroughly described, including the models, datasets, and hyperparameters. As AsymKV is a training-free approach, the compression strategy is clearly outlined, providing sufficient detail for others to replicate the results.
    
    \item {\bf Open access to data and code}
        \item[] Question: Does the paper provide open access to the data and code, with sufficient instructions to faithfully reproduce the main experimental results, as described in supplemental material?
        \item[] Answer: \answerYes{} % Replace by \answerYes{}, \answerNo{}, or \answerNA{}.
        \item[] Justification: Will be added in the supplemental material.

    \item {\bf Experimental setting/details}
        \item[] Question: Does the paper specify all the training and test details (e.g., data splits, hyperparameters, how they were chosen, type of optimizer, etc.) necessary to understand the results?
        \item[] Answer: \answerYes{} % Replace by \answerYes{}, \answerNo{}, or \answerNA{}.
        \item[] Justification: Since AsymKV is training-free, no training details are required. Meanwhile, the paper provides comprehensive inference settings, including compression rates, chunk sizes, and model configurations. Key hyperparameters, such as max\_length and chunk size, are explicitly listed, ensuring clarity.
    
    % TODO
    \item {\bf Experiment statistical significance}
        \item[] Question: Does the paper report error bars suitably and correctly defined or other appropriate information about the statistical significance of the experiments?
        \item[] Answer: \answerYes{} % Replace by \answerYes{}, \answerNo{}, or \answerNA{}.
        \item[] Justification: The paper does not include traditional error bars due to the computational expense of long-context experiments, which allowed for only one run per configuration. However, the statistical significance of results is established through comprehensive evaluations across multiple models (Llama-3.1-8B, Mistral-7B, Qwen2-7B, Llama-2-7B) and various settings. This extensive cross-model validation demonstrates the consistent performance advantages of AsymKV across different settings, providing strong evidence for the robustness of the reported results.
    
    \item {\bf Experiments compute resources}
        \item[] Question: For each experiment, does the paper provide sufficient information on the computer resources (type of compute workers, memory, time of execution) needed to reproduce the experiments?
        \item[] Answer: \answerYes{} % Replace by \answerYes{}, \answerNo{}, or \answerNA{}.
        \item[] Justification: The paper specifies that experiments were run on NVIDIA A100 80GB GPUs and includes details on memory usage in \ref{tab:memory_consumption}.
        
    \item {\bf Code of ethics}
        \item[] Question: Does the research conducted in the paper conform, in every respect, with the NeurIPS Code of Ethics \url{https://neurips.cc/public/EthicsGuidelines}?
        \item[] Answer: \answerYes{} % Replace by \answerYes{}, \answerNo{}, or \answerNA{}.
        \item[] Justification: The research focuses on improving LLM efficiency and does not involve human subjects, sensitive data, or applications with potential harm. It fully aligns with the ethical standards outlined in the NeurIPS Code of Ethics.
    
    % TODO
    \item {\bf Broader impacts}
        \item[] Question: Does the paper discuss both potential positive societal impacts and negative societal impacts of the work performed?
        \item[] Answer: \answerNo{} % Replace by \answerYes{}, \answerNo{}, or \answerNA{}.
        \item[] Justification: This paper is a foundational research and does not directly point to any potential negative social impacts.
        
    \item {\bf Safeguards}
        \item[] Question: Does the paper describe safeguards that have been put in place for responsible release of data or models that have a high risk for misuse (e.g., pretrained language models, image generators, or scraped datasets)?
        \item[] Answer:\answerNA{} % Replace by \answerYes{}, \answerNo{}, or \answerNA{}.
        \item[] Justification: No new high-risk assets released.
    % TODO Longbenchv2、LEval引用
    \item {\bf Licenses for existing assets}
        \item[] Question: Are the creators or original owners of assets (e.g., code, data, models), used in the paper, properly credited and are the license and terms of use explicitly mentioned and properly respected?
        \item[] Answer: \answerYes{} % Replace by \answerYes{}, \answerNo{}, or \answerNA{}.
        \item[] Justification: the paper credits the original sources of datasets and models, mentioning applicable licenses and adhering to their terms of use.
    
    \item {\bf New assets}
        \item[] Question: Are new assets introduced in the paper well documented and is the documentation provided alongside the assets?
        \item[] Answer: \answerYes{} % Replace by \answerYes{}, \answerNo{}, or \answerNA{}.
        \item[] Justification: We release the implementation code of our proposed method to support reproducibility.
        \begin{itemize}
            \item \textbf{Asset Type}: Source code (implementation of the proposed method \textit{AsymKV}).
            \item \textbf{Intended Use:} Research and reproducibility. Enables replication and extension of the reported results. 
            \item \textbf{License:} Apache License 2.0
            \item \textbf{Repository:} \url{https://github.com/the-scale-lab/AsymKV}
            \item \textbf{Documentation:} Usage instructions, environment setup, and experimental commands are provided in the repository \texttt{README.md}.
            \item \textbf{Ethical Considerations:} No human or personal data are involved. All external datasets and models comply with their original licenses.
        \end{itemize}
    
    \item {\bf Crowdsourcing and research with human subjects}
        \item[] Question: For crowdsourcing experiments and research with human subjects, does the paper include the full text of instructions given to participants and screenshots, if applicable, as well as details about compensation (if any)? 
        \item[] Answer: \answerNA{} % Replace by \answerYes{}, \answerNo{}, or \answerNA{}.
        \item[] Justification: No human subjects involved.
    
    \item {\bf Institutional review board (IRB) approvals or equivalent for research with human subjects}
        \item[] Question: Does the paper describe potential risks incurred by study participants, whether such risks were disclosed to the subjects, and whether Institutional Review Board (IRB) approvals (or an equivalent approval/review based on the requirements of your country or institution) were obtained?
        \item[] Answer: \answerNA{} % Replace by \answerYes{}, \answerNo{}, or \answerNA{}.
        \item[] Justification: No human subjects are involved in this research, rendering IRB approvals unnecessary.
    % ToDo 需要确认
    \item {\bf Declaration of LLM usage}
        \item[] Question: Does the paper describe the usage of LLMs if it is an important, original, or non-standard component of the core methods in this research? Note that if the LLM is used only for writing, editing, or formatting purposes and does not impact the core methodology, scientific rigorousness, or originality of the research, declaration is not required.
        %this research? 
        \item[] Answer: \answerYes{} % Replace by \answerYes{}, \answerNo{}, or \answerNA{}.
        \item[] Justification: The paper clearly states that LLMs serve as the base models for evaluating AsymKV. Their usage is standard and well-documented, requiring no further declaration beyond what is provided.

\end{enumerate}
%%%%%%%%%%%%%%%%%%%%%%%%%%%%%%%%%%%%%%%%%%%%%%%%%%%%%%%%%%%%
\newpage
\appendix
\input{analysis}
\input{append}

%%%%%%%%%%%%%%%%%%%%%%%%%%%%%%%%%%%%%%%%%%%%%%%%%%%%%%%%%%%%

\end{document}

%% file: intro.tex
\section{Introduction}

The ability to process long contexts is crucial for Large Language Models (LLMs)
%, as recent studies reveal that precise reasoning and complex task completion heavily depend on extended context windows
~\cite{levy-etal-2024-task,MoonshotAI}. However, processing such long contexts poses significant challenges: pre-trained LLMs face both architectural and computational constraints in handling extended contexts. In particular, as the context length increases, the complexity of attention mechanisms increases quadratically ($O(n^2)$), while storage overhead increases linearly ($O(n)$)~\cite{dao2022flashattention}. 

Various approaches have been proposed to address this challenge, with KV cache compression emerging as a promising direction~\cite{li2024survey}. These methods aim to compress the KV cache while preserving essential information for maintaining model performance. A straightforward strategy is to keep tokens with high historical importance (e.g., attention scores~\cite{NEURIPS2024_28ab4182,liu2024scissorhands,oren-etal-2024-transformers,zhang2023h2o}). This approach leverages the observation that attention weights exhibit significant variation across different tokens. Another line of work attempts to identify more general token importance rather than the history information~\cite{xiao2023streamingllm,wang2024modeltellsmergeadaptive,chen2024sepllm}. %For example, StreamingLLM~\cite{xiao2023streamingllm} discovers that initial and recent tokens significantly impact subsequent predictions. 
However, these approaches share a fundamental limitation: they fail to capture certain tokens that are less important in the histroy but suddenly become critical for subsequent predictions. 

To address the information loss caused by directly discarding tokens, cache merging methods have been proposed to merge multiple tokens into fewer representations rather than hard pruning, thereby preserving more information~\cite{zhangcam,wan2025textdtexto,wang2024modeltellsmergeadaptive}. These merging approaches implicitly assume that certain redundancies or patterns exist in the KV cache. This raises a new fundamental question: {\it what specific characteristics of LLMs lead to these redundancies and make cache merging feasible?} We answer this question by identifying the key-value asymmetry in LLM attention mechanisms.

\begin{figure*}[t]  
    \centering  
    \begin{subfigure}[b]{0.23\textwidth}  
        \centering  
        \includegraphics[width=\textwidth]{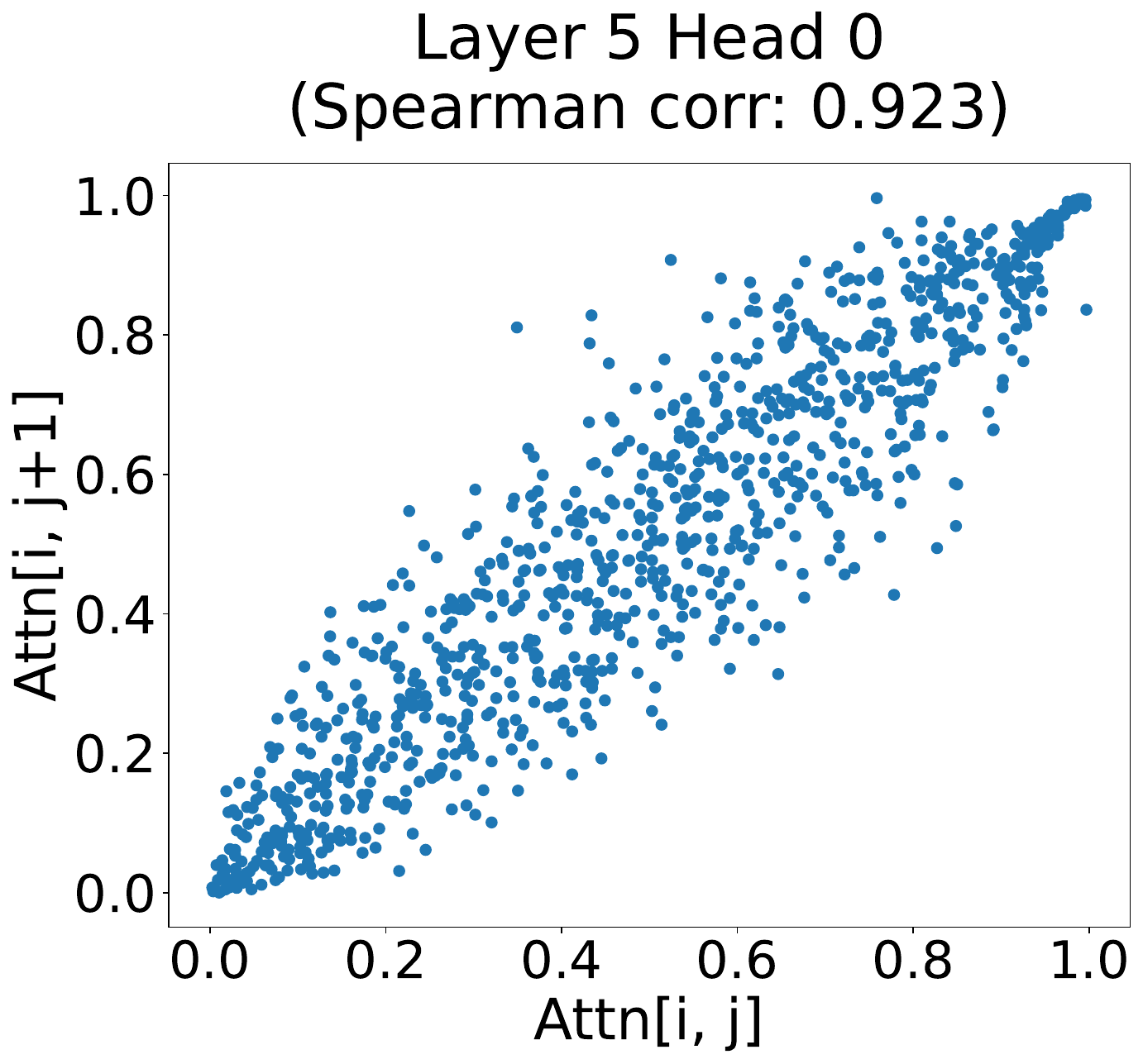}  
        \caption{Scatter plot of {\bf key} similarity (attention) weight percentile ranks between adjacent positions}  
        \label{fig:scater_attn}  
    \end{subfigure}  
    \begin{subfigure}[b]{0.23\textwidth}  
        \centering  
        \includegraphics[width=\textwidth]{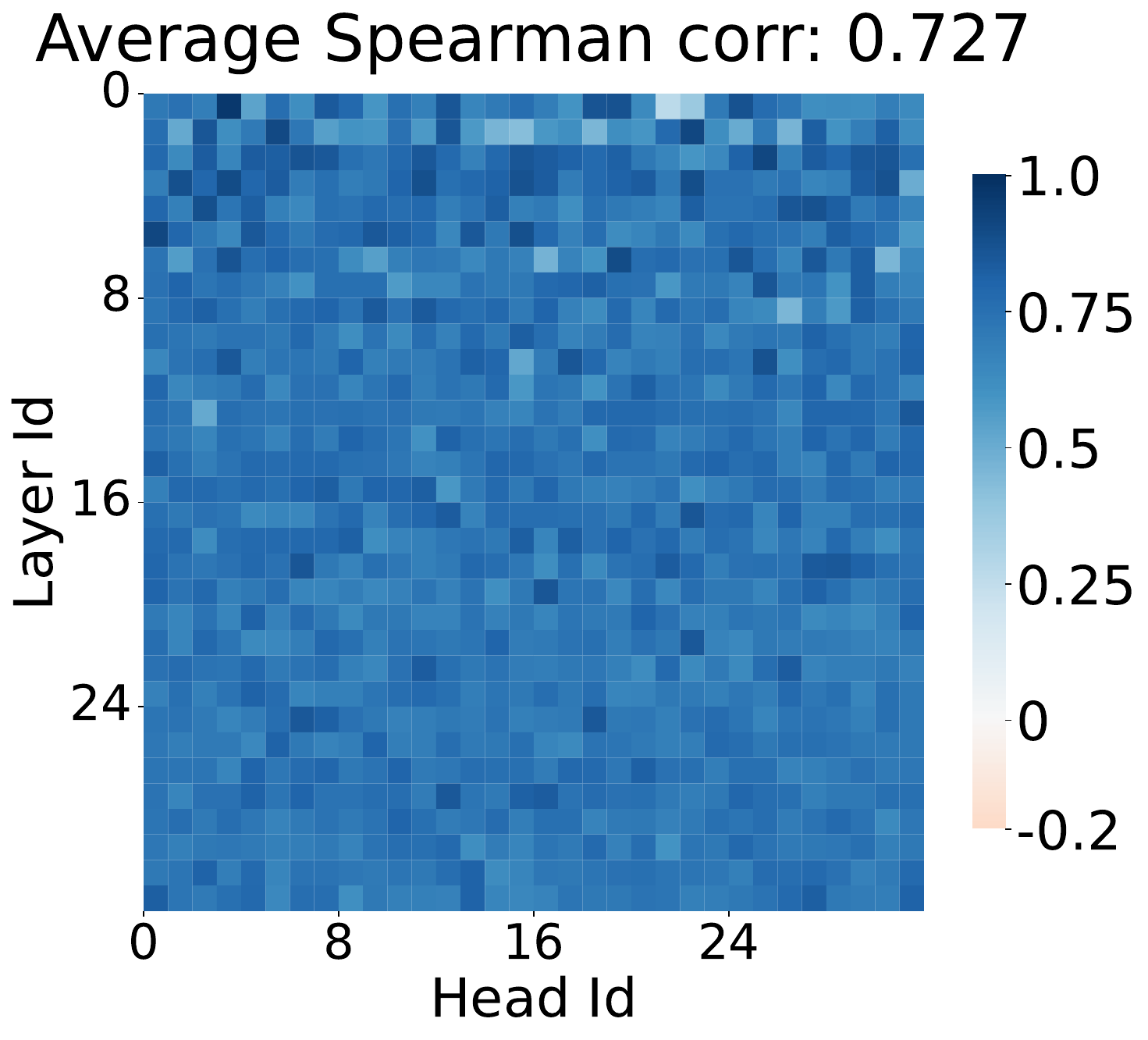}  
        \caption{Spearman correlation map of adjacent {\bf attention weights} \newline}
        \label{fig:heatmap_attn_llama2}  
    \end{subfigure}  
    \begin{subfigure}[b]{0.23\textwidth}  
        \centering  
        \includegraphics[width=\textwidth]{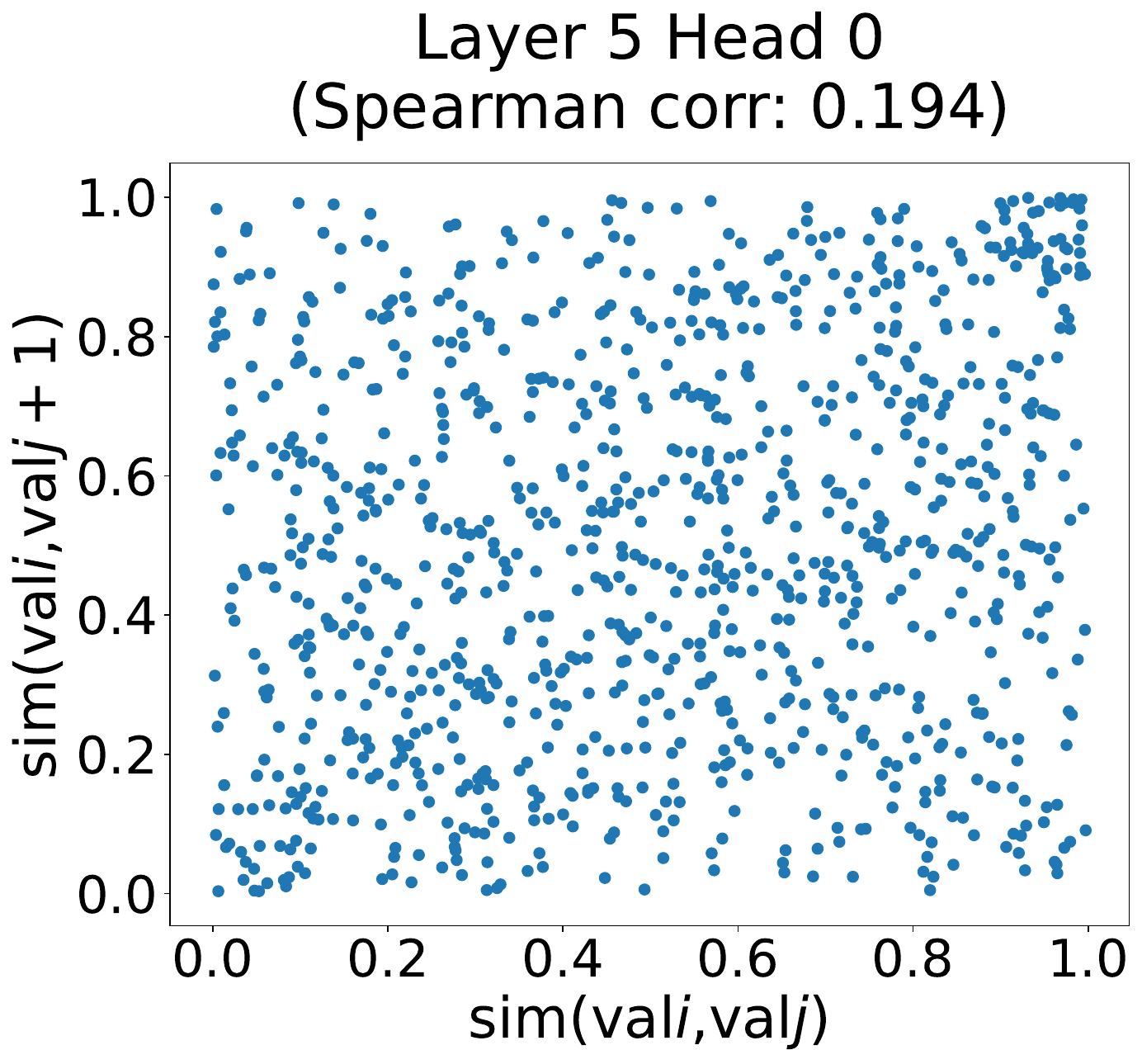}  
        \caption{Scatter plot of {\bf value} similarity percentile ranks between adjacent positions}  
        \label{fig:spersonr_value}  
    \end{subfigure}  
    \begin{subfigure}[b]{0.23\textwidth}  
        \centering  
        \includegraphics[width=\textwidth]{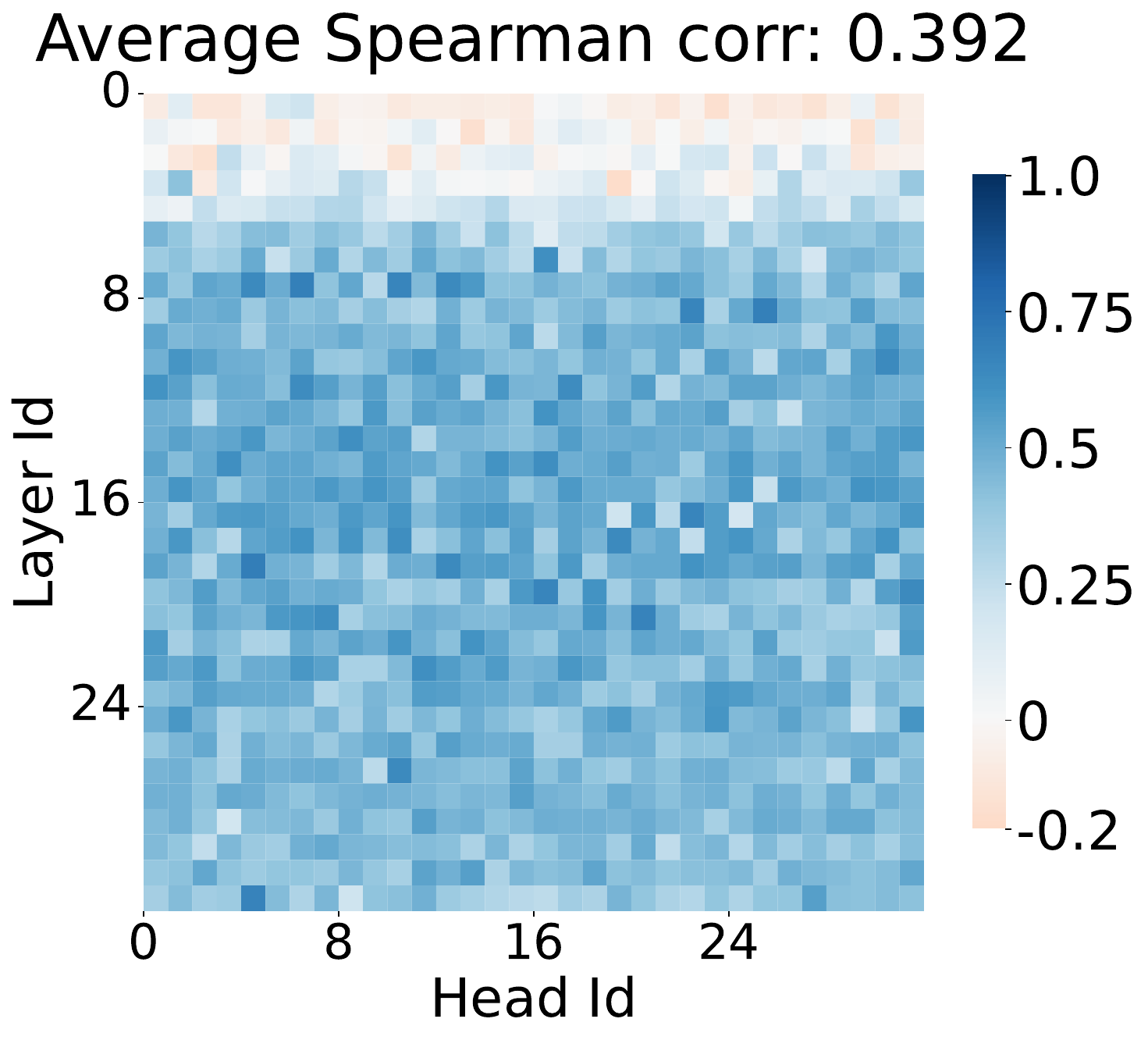}  
        \caption{Spearman correlation map of adjacent {\bf value} similarity \newline}
        \label{fig:heatmap-value}  
    \end{subfigure}  
    \caption{Contrasting distributions of {\bf local homogeneity in attentions (keys) versus local heterogeneity in values}. Statistics are from Llama-2-7b-chat on the ShareGPT dataset. (a-b) demonstrate strong positive correlations between adjacent attention percentile ranks (normalized to [0,1], where 1 indicates highest attention) across all layers and heads, supporting the local homogeneity hypothesis for keys. (c-d) reveal weak or negative correlations between adjacent value similarity percentile ranks, computed from $\text{sim}(\text{val}_i,\text{val}_j)$, indicating distinct heterogeneity in values. The similarity is measured by cosine. This fundamental difference between keys and values suggests the need for separate compression strategies.}  
    \label{fig:all-heatmaps}  
\end{figure*}

\paragraph*{Local Key-Value Asymmetry} Through extensive empirical analysis, we reveal a fundamental pattern in attention distributions: the \textit{homogeneity of local keys}. Specifically, we observe that {\bf adjacent tokens consistently receive similar attention weights} - when a query assigns high attention to position $j$, the neighboring position ($j+1$) typically receives comparable attention weight (Fig.~\ref{fig:scater_attn}). This pattern shows remarkable consistency across all layers and attention heads, with an average Spearman correlation coefficient of 0.727 (Fig.~\ref{fig:heatmap_attn_llama2}). This consistent local attention pattern, arising from query-key interactions, suggests an underlying \textit{homogeneity of local keys} - adjacent keys must share certain structural properties to produce such stable attention patterns. Such key homogeneity naturally emerges from language structure, where adjacent words form coherent semantic units and contribute collectively to meaning representation. 

%This discovery provides a new insight for KV cache compression. Rather than focusing on individual token importance, our findings highlight the critical role of token relationships, shifting the question from ``which tokens are important'' to ``how tokens interact with their neighbors''. This structural understanding subsumes previous token-centric patterns as special cases - for instance, the effectiveness of retaining initial and recent tokens in StreamingLLM can be explained by the homogeneous attention patterns within these regions. More importantly, this perspective enables us to identify and preserve crucial tokens that might be overlooked by individual importance measures.  

The observed homogeneity of adjacent keys provides evidence for merging neighboring tokens, offering a principled explanation to recent token merging approaches. Specifically, when multiple adjacent keys exhibit high homogeneity, computing and storing only one key for the merged representation effectively approximate the original attention output, leading to both computational and memory efficiency.

However, our analysis reveals a striking asymmetry: while keys exhibit strong local homogeneity, adjacent values demonstrate markedly distinct \textit{heterogeneous distributions}. As shown in Fig.~\ref{fig:spersonr_value} and Fig.~\ref{fig:heatmap-value}, when switching from keys to value similarities, adjacent value vectors ($\mathbf{v}_i$ and $\mathbf{v}_{i+1}$) often show much lower or even negative correlations in some layers. 

%This key-value asymmetry reveals critical limitations in existing methods: cache compression approaches~\cite{li2024snapkv,liu2024scissorhands,zhang2023h2o,xiao2023streamingllm} discard keys and their corresponding values together, failing to recognize that homogeneous keys might correspond to heterogeneous values that carry distinct information; similarly, cache merging methods~\cite{zhangcam,wan2024d2o,wang2024model} apply identical merging strategies to both keys and values, overlooking their fundamentally different distributional characteristics.

This local key-value asymmetry reveals critical limitations in existing methods: cache merging methods~\cite{zhangcam,wan2025textdtexto} apply identical merging strategies to both keys and values, overlooking their fundamentally different distributional characteristics. More studies of this phenomenon will be discussed in \S~\ref{sec:analysis}.

\paragraph*{Training-free Asymmetry Modeling} 
Based on the above analysis, the key challenge in cache merging lies in modeling heterogeneous values. Fortunately, through careful examine of attention mechanisms' mathematical structure, we discover an elegant solution to this value heterogeneity. We develop a mathematically proven value representation scheme that guarantees lossless attention computation after merging adjacent keys. Notably, our method remains distribution-agnostic, making it inherently robust to value heterogeneity.

Building on the key-value asymmetry and the property for values, we propose AsymKV, a novel training-free cache merging method for efficient long-context modeling. Our key insight is to shift the information loss from heterogeneous values to homogeneous keys during merging, thereby minimizing overall loss. Extensive experiments demonstrate that our method consistently outperforms existing long-context methods across various tasks and base models. On LLaMA3.1-8B, AsymKV achieves an average score of 43.95 on LongBench, surpassing H$_2$O~\cite{zhang2023h2o} (38.89) by a significant margin. These results demonstrate AsymKV's effectiveness in extending LLMs' context handling capabilities without additional training.

{\bf Our Contributions:} The key contributions of this work are threefold. First, we reveal a contrasting, yet previously overlooked asymmetry of local keys and values in LLM attention mechanisms. Second, based on this asymmetric property, we propose a novel training-free compression framework that combines homogeneity-based key merging with a mathematically proven lossless value representation. Third, we demonstrate through extensive experiments that our method consistently outperforms existing long-context methods across various tasks and base models.

%% file: related.tex
\section{Related Work}

%Enhancing the long-context capabilities of language models has emerged as a critical problem~\cite{yunpeng2023advancing,xindi2024beyond,jiaqi2023loogle}. Various methods have been proposed. We discuss three techniques that are more related to our study below.

%\textbf{Linear Sequence Modeling Architectures} To tackle the inefficiency in processing long sequences, researchers have introduced several novel sequence modeling architectures with linear complexity. For instance, \textit{Mamba} \cite{mamba2023} significantly improves the efficiency of long sequence processing by redesigning the flow of information. The core idea of such architectures is to optimize the representation of sequences, enabling the model to better capture long-range dependencies while maintaining computational efficiency.  However, these new information flow propagation methods may lead to a decrease in the model's expressive power, and therefore have not been widely used in pre-trained language models.

\textbf{KV Cache Pruning} Recent research focuses on compressing the KV cache through selective token retention and importance-based pruning. H$_2$O~\cite{zhang2023h2o} introduces the concept of ``Heavy Hitters'' - tokens that contribute significantly to attention scores - and develops a theoretically-grounded eviction policy. Building on this idea, RoCo~\cite{ren2024efficacy} improves the robustness of cache compression by considering both temporal attention scores and stability measures. More recent works like SnapKV~\cite{NEURIPS2024_28ab4182} and Scissorhands~\cite{liu2024scissorhands} leverage the persistence of token importance across generation steps, while \citep{devoto-etal-2024-simple} demonstrates that $L_2$ norm-based compression can achieve competitive results with a simpler implementation. However, these compression methods face a fundamental challenge: they rely heavily on token-centric measures (e.g., attention scores or norm values) to determine which tokens to retain, potentially discarding tokens that suddenly become crucial for future predictions.

\textbf{KV Cache Merging}   
Another line of works have explored merging similar KV cache positions to reduce memory footprint during inference. CaM~\cite{zhangcam} proposes an adaptive merging strategy guided by attention scores, while D$_2$O~\cite{wan2025textdtexto} introduces a two-level discriminative approach considering both layer-wise patterns and token similarities. KVMerger~\cite{wang2024modeltellsmergeadaptive} adaptively constructs the KV cache by analyzing the intrinsic structure of attention modules. However, a fundamental limitation of these approaches is their uniform treatment of keys and values during merging despite their distinct distributional characteristics. As discovered in the introduction, this oversight is particularly problematic given the inherent heterogeneity of value vectors, which, unlike keys, often exhibit significant variations even between adjacent positions.

\textbf{Context Segmentation and Sliding} One polular variant of KV cache compression leverages context segmentation and sliding. These approaches stem from \textit{StreamingLLM}~\cite{xiao2023streamingllm}, which discovered that initial tokens and recent tokens are more important than other middle tokens in the attention. Therefore, it only keeps such tokens. \textit{LongCache}~\cite{liu2024farewell} expands StreamingLLM by keep critic middle tokens. These tokens are identified via the historical attention weights. \textit{SirLLM}~\cite{yao-etal-2024-sirllm} uses token entropy to identify and keeps critic middle tokens. These segmentation-based methods require minimal KV cache operations, making them computationally efficient. However, these compression strategies essentially involve directly discarding tokens with lower weights, which results in significant information loss when these tokens suddenly become critical for future predictions.

%% file: method.tex
\section{Proposed Method}

%Building on our analysis of key-value asymmetry, we propose AsymKV, a novel attention mechanism that efficiently handles homogeneous keys while preserving the distinct characteristics of heterogeneous values in the KV cache. 

\begin{figure}[t]
    \centering
    \includegraphics[width=\linewidth]{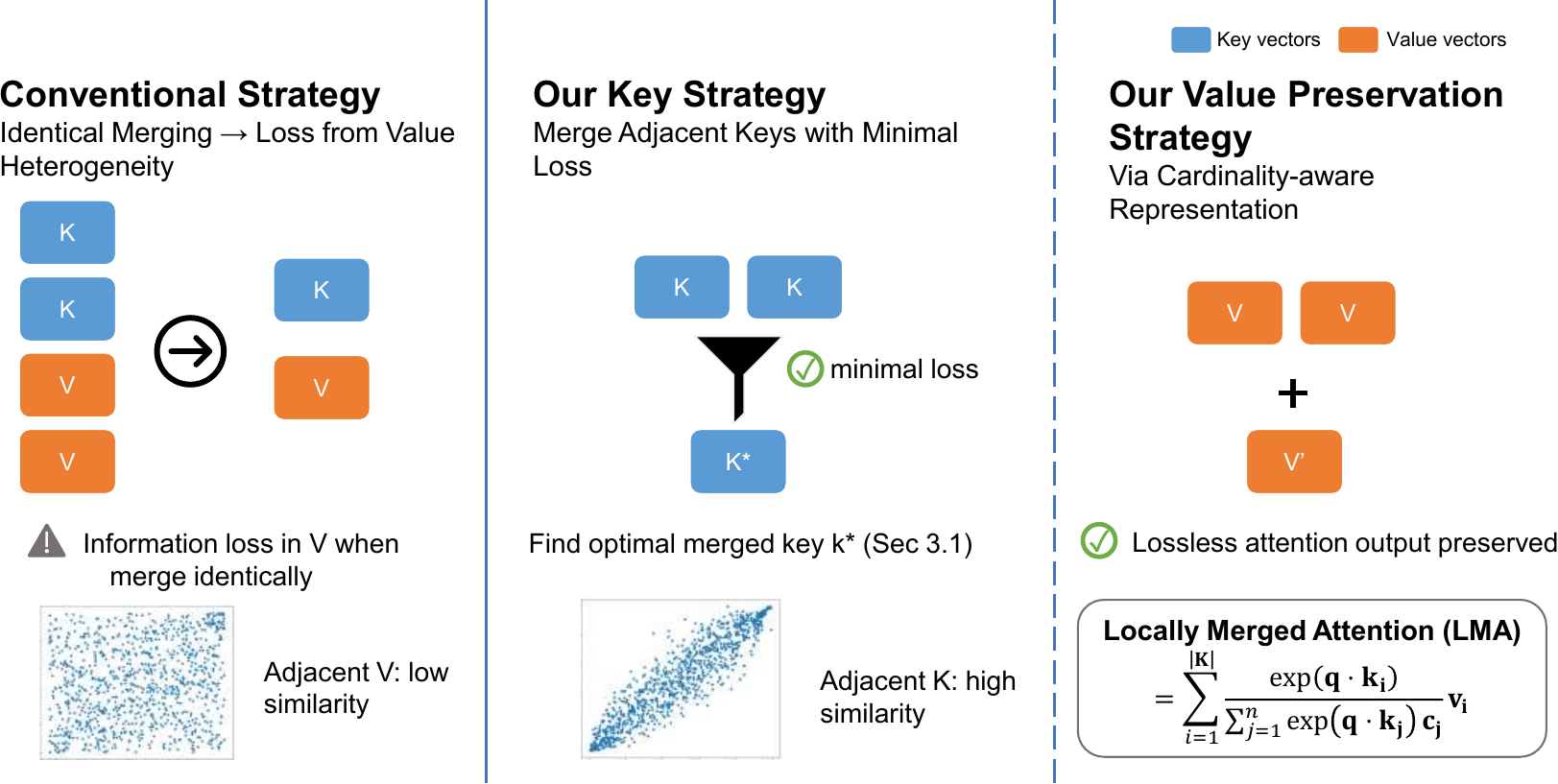}
    \caption{Illustration of our AsymKV mechanism. 
    Left: Conventional approaches that uniformly merge both keys and values lead to information loss. Middle: We merge adjacent homogeneous keys for minimal loss. Right: We preserve their heterogeneous values through cardinality-aware normalization.}
    \label{fig:method}
\end{figure}

Building on the key-value asymmetry, we propose AsymKV. Our main idea is to shift the information loss from heterogeneous values to homogeneous keys during merging, thereby minimizing the loss. We show the key intuition and framework of AsymKV in Figure~\ref{fig:method}. (Left) Previous approaches that apply identical merging strategies to both keys and values suffer from significant information loss, especially considering the heterogeneous nature of values. In contrast, we leverage the asymmetry between keys and values in the attention mechanism. Our method compresses keys with minimal information loss (\S~\ref{sec:method:key}) due to their local homogeneity nature, while preserving the distinct characteristics of heterogeneous values through cardinality-aware normalization (\S~\ref{sec:method:value}).

%enabling efficient and effective KV cache merging  (\S~\ref{sec:method:parallel}).

%In this section, we start with compressing one pair of adjacent tokens. We will demonstrate how to find the optimal compressed key in \S~\ref{sec:method:key}, how to introduce a cardinality-aware normalization for a lossless value pair compression in \S~\ref{sec:method:value}. Then we apply the compression in the next-token prediction paradigm. We show how to compress multiple adjacent tokens in parallel for efficient generation in \S~\ref{sec:method:parallel}.

%This section presents our method in three parts: \S~\ref{sec:method:key}: A theoretical framework for deriving optimal compressed keys from adjacent token pairs; \S~\ref{sec:method:value}: A cardinality-aware normalization scheme that enables lossless value compression while preserving heterogeneous information; \S~\ref{sec:method:parallel}: An efficient parallel compression strategy for handling multiple adjacent tokens during generation.

\subsection{Homogeneous Key Merging}
\label{sec:method:key}
Our primary goal of adjacent token compression is to convert the original $n$ tokens in the KV cache into $n-1$ by merging a pair of adjacent positions $m, m+1$.

First, consider the key merging. Based on our observation of adjacent key homogeneity, we can merge adjacent keys without significantly affecting model performance. Given key vectors $\mathbf{K} = [\mathbf{k}_1, \mathbf{k}_2, \dots, \mathbf{k}_n]$ from the KV cache and a pair of adjacent positions $m, m+1$ to be merged, let $\mathcal{L}(\mathbf{K})$ denote the language modeling loss. After compressing $\mathbf{k}_m, \mathbf{k}_{m+1}$ into one embedding $\mathbf{k}$, we denote the new loss as: 
\begin{equation}  
\label{eq:mismatch}
\mathcal{L}([\mathbf{K}_{<m}, \mathbf{k}, \mathbf{K}_{>m+1}]) 
\end{equation}  
where $\mathbf{K}_{<i}$ denotes the sequence $[\mathbf{k}_1, \dots, \mathbf{k}_{i-1}]$, and $\mathbf{K}_{>i}$ denotes $[\mathbf{k}_{i+1}, \dots, \mathbf{k}_n]$.
Our objective of optimal key compression is to find $\mathbf{k}$ that minimizes the information loss.  

Due to the dimensional mismatch between the original $\mathbf{K}$ and $[\mathbf{K}_{<m}, \mathbf{k}, \mathbf{K}_{>m+1}]$ in Eq.~\eqref{eq:mismatch} ($n \times d \to (n-1) \times d$), our cache merging takes two steps:  
{\bf 1.} Find a pair of identical embeddings $(\mathbf{k}^*, \mathbf{k}^*)$ to replace $(\mathbf{k}_m, \mathbf{k}_{m+1})$ while preserving dimensionality, which is mathematically tractable.  
{\bf 2.} Leverage attention properties to merge the two tokens. 

We first find optimal embeddings $\mathbf{k}$ that minimize $\mathcal{L}([\mathbf{K}_{<m}, \mathbf{k}, \mathbf{k}, \mathbf{K}_{>m+1}])$ while keeping the dimensionality. For simplification, we denote it as $\mathcal{L}(\mathbf{k}, \mathbf{k})$, and the optimal $\mathbf{k}$ as $\mathbf{k}^*$: $\mathbf{k}^* = \mathop{\arg\min}_\mathbf{k} \mathcal{L}(\mathbf{k}, \mathbf{k})$.

We approach this optimization problem using a Newton-like method. By applying a second-order Taylor expansion of $\mathcal{L}(\mathbf{x}, \mathbf{y})$ around $(\mathbf{k}_m, \mathbf{k}_{m+1})$:
\begin{equation}  
\begin{aligned}  
\mathcal{L}(\mathbf{x}, \mathbf{y}) &\approx \mathcal{L}(\mathbf{k}_m, \mathbf{k}_{m+1}) +   
\nabla \mathcal{L}(\mathbf{k}_m, \mathbf{k}_{m+1})^\top 
\begin{bmatrix}  
\mathbf{x} - \mathbf{k}_m \\
\mathbf{y} - \mathbf{k}_{m+1}  
\end{bmatrix} 
+ \frac{1}{2}  
\begin{bmatrix}  
\mathbf{x} - \mathbf{k}_m \\
\mathbf{y} - \mathbf{k}_{m+1}  
\end{bmatrix}^\top  
\mathbf{H}  
\begin{bmatrix}  
\mathbf{x} - \mathbf{k}_m \\
\mathbf{y} - \mathbf{k}_{m+1}  
\end{bmatrix}  
\end{aligned}  
\end{equation}  
where $\mathbf{H}$ is the Hessian matrix at $(\mathbf{k}_m,\mathbf{k}_{m+1})$. We denote the Hessian matrix $\mathbf{H}$ as:  
\begin{equation}  
\mathbf{H} = \begin{bmatrix}  
\mathbf{H}^{11} & \mathbf{H}^{12} \\
\mathbf{H}^{21} & \mathbf{H}^{22}  
\end{bmatrix}  
\end{equation}  
Each submatrix $\mathbf{H}^{ab}$ is a $d \times d$ matrix. 
To minimize $\mathcal{L}(\mathbf{k}, \mathbf{k})$, we set $\mathbf{x} = \mathbf{k},\; \mathbf{y} = \mathbf{k}$ and substitute into our quadratic approximation:
\begin{equation}
\begin{aligned}
\label{eq:method:optimal_compress:first_step:loss_expansion}
& \mathcal{L}(\mathbf{k}, \mathbf{k}) \approx \mathcal{L}(\mathbf{k}_m, \mathbf{k}_{m+1}) + \nabla \mathcal{L}(\mathbf{k}_m, \mathbf{k}_{m+1})^\top
\begin{pmatrix}
\mathbf{k} - \mathbf{k}_m \\
\mathbf{k} - \mathbf{k}_{m+1}
\end{pmatrix}
+ \frac{1}{2} 
\begin{pmatrix}
\mathbf{k} - \mathbf{k}_m \\
\mathbf{k} - \mathbf{k}_{m+1}
\end{pmatrix}^\top
\mathbf{H}
\begin{pmatrix}
\mathbf{k} - \mathbf{k}_m \\
\mathbf{k} - \mathbf{k}_{m+1}
\end{pmatrix}
\end{aligned}
\end{equation}

Following the Newton method, we find the critical point by setting the gradient of this quadratic approximation to zero. This yields the optimal solution (details in Appendix~\ref {sec:append:solving_for_the_optimal_embedding_vector}):
\begin{equation}  
\begin{aligned}  
\label{eq:optimal_k}  
\mathbf{k}^* = &(\mathbf{H}^{11} + 2\mathbf{H}^{12} + \mathbf{H}^{22})^{-1}[  
\mathbf{H}^{11}\mathbf{k}_m + \mathbf{H}^{12}(\mathbf{k}_m + \mathbf{k}_{m+1}) 
+ \mathbf{H}^{22}\mathbf{k}_{m+1} - (\mathbf{g}_m + \mathbf{g}_{m+1})]  
\end{aligned}  
\end{equation}  
where $\mathbf{g}_m = \nabla_{\mathbf{k}_m} \mathcal{L}$ and $\mathbf{g}_{m+1} = \nabla_{\mathbf{k}_{m+1}} \mathcal{L}$. 

To efficiently compute the Hessian matrix $\mathbf{H}$, we use the Fisher information matrix as an approximation, which is a common technique in second-order optimization methods. Following the approach in neural network pruning~\citep{lecun1989optimal}, we assume that the interactions between parameters are negligible and approximate the Fisher information matrix as a diagonal matrix. The diagonal elements can be efficiently computed using their gradients:
\begin{equation}
\mathbf{H}_{ii} = F_{ii} = \nabla \mathcal{L}(\mathbf{k}_m, \mathbf{k}_{m+1})_i^2
\end{equation}

In our empirical analysis, we discover that the magnitude of gradient terms $\mathbf{g}_m$ and $\mathbf{g}_{m+1}$ can exceed that of $\mathbf{H}$ by six orders of magnitude in Eq.~\eqref{eq:optimal_k}. This is a known issue in Newton-like methods when the function is far from its minimum or when the curvature is very small. To stabilize the optimization and maintain $\mathbf{k}^*$ as a valid key, we adopt a modified Newton approach by dropping the gradient terms from Eq.~\eqref{eq:optimal_k}, effectively using only the curvature information to guide our solution.

{\bf Compression Position Selection} To minimize information loss during merging, we select positions $m, m+1$ with the lowest sum of attention scores, where positions receiving minimal attention have the least impact on the model's attention mechanism.

\iffalse
For each layer, we compute a token importance score by aggregating attention weights across all heads and queries:  
\begin{equation}  
    \label{eq:method:position:score}  
\text{score}_i = \sum_{h=1}^{H} \sum_{j=1}^{n-1} (\text{Attn}^{(h)}_{j,i} + \text{Attn}^{(h)}_{j,i+1} ) 
\end{equation}  
where $H$ is the number of attention heads and $\text{Attn}^{(h)}_{j,i}$ represents the attention weight from query $j$ to key $i$ in head $h$. We select the position $m$ with the lowest score in each layer independently, allowing for layer-specific compression that better preserves the model's predictive capabilities.  
\fi

\subsection{Cardinality Normalization for Lossless Value Merging}
\label{sec:method:value}

After replacing adjacent keys with identical embeddings, two challenges remain: (1) how to reduce input tokens $(\mathbf{k}^*, \mathbf{k}^*) \rightarrow \mathbf{k}^*$ to improve computing efficiency; (2) how to merge their corresponding values. We elaborate how to extend the attention mechanism to address both challenges while maintaining output equivalence.

In the original attention mechanism, the output for query $\mathbf{q}$ is:  
\begin{equation}  
\text{Attention}(\mathbf{q}, \mathbf{K}, \mathbf{V}) = \sum_{i=1}^{|\mathbf{K}|} \frac{\exp(\mathbf{q} \cdot \mathbf{k}_i)}{\sum_{j=1}^{|\mathbf{K}|} \exp(\mathbf{q} \cdot \mathbf{k}_j)} \mathbf{v}_i  
\end{equation}  

After key compression where $\mathbf{k}_m = \mathbf{k}_{m+1} = \mathbf{k}^*$ in \S~\ref{sec:method:key}, we have:  
\begin{equation}  
\label{eq:attn_identical}
\begin{aligned}  
\text{Attention}&(\mathbf{q}, \mathbf{K}, \mathbf{V}) = \sum_{i \in [1,n]\setminus\{m,m+1\}} \frac{\exp(\mathbf{q} \cdot \mathbf{k}_i)}{\sum_{j=1}^{|\mathbf{K}|} \exp(\mathbf{q} \cdot \mathbf{k}_j)} \mathbf{v}_i   
+ \underbrace{\frac{\exp(\mathbf{q} \cdot \mathbf{k}^*)}{\sum_{j=1}^n \exp(\mathbf{q} \cdot \mathbf{k}_j)} (\mathbf{v}_m + \mathbf{v}_{m+1})}_{\mbox{\begin{tabular}{c} \normalsize attention to one merged KV pair \\ \normalsize (key=$\mathbf{k}^*$, value=$\mathbf{v}_m+\mathbf{v}_{m+1}$) \end{tabular}}}  
\end{aligned}  
\end{equation}  

Examining Eq.~\eqref{eq:attn_identical}, we observe a key insight: after converting both $\mathbf{k}_m, \mathbf{k}_{m+1}$ to $\mathbf{k}^*$, the attention output for two original tokens $m$ and $m+1$ is mathematically equivalent to the attention output for a single compressed token with key $\mathbf{k}^*$ and value $(\mathbf{v}_m + \mathbf{v}_{m+1})$.

{\bf (Locally Merged Attention)} The insight above naturally suggests an alternated attention mechanism for merged tokens, which we denote as Locally Merged Attention (LMA):
\begin{equation}  
\label{eq:hla}  
\text{LMA}(\mathbf{q},\mathbf{K},\mathbf{V},\mathbf{C}) = \sum_{i=1}^{|\mathbf{K}|} \frac{\exp(\mathbf{q} \cdot \mathbf{k}_i)}{\sum_{j=1}^n \exp(\mathbf{q} \cdot \mathbf{k}_j) \mathbf{c}_j} \mathbf{v}_i   
\end{equation}  
where $\mathbf{c}_i$ indicates the number of original tokens represented by the $i$-th compressed token. The cardinality vector $\mathbf{c}$ is designed to maintain the denominator in Eq.~\eqref{eq:attn_identical}, ensuring mathematical equivalence between the original and compressed attention mechanisms. Initially, $\mathbf{c}_i=1$ for all tokens. After merging positions $m$ and $m+1$, we update $\mathbf{C}$ as $[\mathbf{C}_{<m}, \mathbf{c}_m + \mathbf{c}_{m+1}, \mathbf{C}_{>m+1}]$, ensuring the denominator in our attention calculation remains equivalent to the original uncompressed attention.

{\bf (Equivalence)} Using LMA, the following equivalence holds:
\begin{equation}  
\text{Attention}(\mathbf{q},\mathbf{K},\mathbf{V}) = \text{LMA}(\mathbf{q},\mathbf{K}',\mathbf{V}',\mathbf{C})  
\end{equation}  
where $\mathbf{K},\mathbf{V}$ are KV caches for $n$ tokens while $\mathbf{K}',\mathbf{V}'$ are for $n-1$ tokens:
\begin{equation}  
\begin{aligned}  
\mathbf{K}' &= [\mathbf{K}_{<m}, \mathbf{k}^*, \mathbf{K}_{>m+1}] \\
\mathbf{V}' &= [\mathbf{V}_{<m}, \mathbf{v}_m + \mathbf{v}_{m+1}, \mathbf{V}_{>m+1}]  
\end{aligned}  
\end{equation}  
This equivalence reveals an elegant characteristic of attention mechanisms: they permit lossless compression of values through simple vector addition. The only cost is to store $\mathbf{C}$ with $n$ integers. This property is particularly powerful as it enables preservation of attention outputs while reducing sequence length, effectively solving the quadratic complexity challenge of long-sequence processing. 

%In \S~\ref{sec:method:parallel}, we will show how to iteratively apply this compression in the next-token prediction process to achieve both storage and computational efficiency.

\subsection{Efficient Implementations for Long-Text Generation}
\label{sec:method:parallel}

\subsubsection{Time-Efficiency by Chunk-wise Parallel Compression}
%While the single-pair compression method described above is theoretically sound, the token-by-token compression approach introduces significant computational overhead. Therefore, 
We propose a chunk-based parallel compression method for efficient long-text generation. The process predicts the next token using HLA in place of the original attention, without requiring any model fine-tuning.
%\textbf{Generation Step:} Predicts the next token using HLA in place of the original attention, without requiring any model fine-tuning. The KV cache is utilized for efficient prediction.  
%\textbf{Compression Step:} 
More specifically, when the context length reaches \texttt{max\_length}, after every \texttt{chunk\_size} new tokens, we compress $\texttt{max\_length}+\texttt{chunk}$ tokens into $\texttt{max\_length}$ tokens in parallel by: 
{\bf 1.} Identifying $\texttt{chunk}$ pairs of adjacent tokens with lowest attention scores.  
{\bf 2.} Computing optimal compression according to Eq.~\eqref{eq:optimal_k}. 
{\bf 3.} Merging keys, values and cardinalities.

Since the merge operation is performed only once every \texttt{chunk\_size} tokens (e.g., 512 tokens), its computational overhead is minimal relative to the overall inference process. The compression step requires only a single backward pass to compute the Hessian matrices for all candidate compression positions, followed by parallel execution of the optimal compression operations. This design ensures that AsymKV maintains inference efficiency. More experimental results are shown in \S~\ref{sec:exp:efficiency}.

\subsubsection{Memory-Efficiency by Selective Gradient Computation}
% Our method requires gradient computation (Eq.~\eqref{eq:optimal_k}), which might raise concerns about increased memory usage. However, AsymKV still maintains memory efficiency compared to other approaches. Unlike typical backpropagation that computes gradients for all parameters, we only compute gradients for {\bf key embeddings within the current chunk, not for all model parameters}. This greatly reduces both memory and computational overhead. Detailed memory statistics are provided in \S~\ref{sec:exp:efficiency}.
Our method requires gradient computation (Eq.~\eqref{eq:optimal_k}) which might raise  concerns about increased memory usage. However, AsymKV still maintains  memory efficiency compared to other approaches. Unlike typical  backpropagation that computes gradients for all parameters, we only  compute gradients for {\bf key embeddings within the current chunk, not for  all model parameters}.

To elaborate further from a quantitative perspective, this selective gradient computation yields a gradient tensor of size approximately \(c \times d\), where \(c\) denotes the chunk size and \(d\) is the dimension per token. In contrast, a standard forward pass (and its associated gradient computation) requires storing the full model parameters (with size \(O(p)\), where \(p\) is the total number of parameters) along with the KV cache states, sized at \(2 \times l \times d\) (where \(l\) represents the maximum sequence length). Thus, the additional memory overhead from our selective gradients is on the order of \(O(cd)\), which is significantly smaller than the baseline’s \(O(2ld + p)\) when \(c \ll l\) (a typical scenario in chunked processing). This avoids unnecessary gradient computations across the entire sequence, ensuring that both memory and computational overhead are greatly reduced. Detailed memory statistics are provided in \S~\ref{sec:exp:efficiency}.

%% file: exp.tex
\section{Experiments}

\begin{table*}[!htbp]
\small
\centering
\caption{Performance on LongBench. AsymKV outperforms its baselines on most settings.}
\begin{adjustbox}{max width=\textwidth}
\begin{tabular}{lccccccc}
\toprule
\textbf{} & \makecell{\textbf{Single-Doc}} & \makecell{\textbf{Multi-Doc}} & \makecell{\textbf{Sum}} & \makecell{\textbf{Few-shot} } & \makecell{\textbf{Synthetic}} & \makecell{\textbf{Code}} & \textbf{Avg.} \\
\midrule
\multicolumn{8}{c}{\textbf{Llama2-7B-chat}} \\
\midrule
Full Context & 25.80 & 21.47 & 24.62 & 62.86 & 4.96 & 48.90 & 32.00 \\
StreamingLLM & 19.29 & 21.05 & 23.15 & 60.85 & 1.81 & 48.58 & 29.61 \\
LongCache & 19.73 & 20.06 & 23.19 & 61.26 & 2.24 & 49.05 & 29.71 \\
H$_2$O & 19.92 & \textbf{25.64 }& 23.85 & 61.37 & 4.27 & 50.28 & 31.34 \\
LLMLingua-2 &21.47 &23.29 &23.53 &33.23&	6.17 &35.12 &24.20 \\
CaM & 19.53 & 20.64 & 22.67 & 61.81 & 4.18 & 48.53 & 30.15 \\

\cellcolor{highlight}{\textbf{AsymKV}} & \cellcolor{highlight}{\textbf{24.63}} & \cellcolor{highlight}{24.15} & \cellcolor{highlight}{\textbf{24.22}} & \cellcolor{highlight}{\textbf{62.11}} & \cellcolor{highlight}{\textbf{10.18}} & \cellcolor{highlight}{\textbf{52.16}} & \cellcolor{highlight}{\textbf{33.12}} \\
\midrule
\multicolumn{8}{c}{\textbf{Llama3.1-8B-Instruct}} \\
\midrule
Full Context & 43.73 & 44.49 & 29.12 & 69.36 & 53.56 & 52.94 & 60.21 \\
StreamingLLM & 28.15 & 27.19 & 25.15 & 63.17 & 16.33 & 54.02 & 35.73 \\
LongCache & 28.98 & 27.84 & 25.35 & 64.73 & 19.68 & 53.60 & 36.70 \\
H$_2$O & 33.30 & 34.43 & 26.60 & {\bf 66.23} & 14.75 & \textbf{55.56} & 38.89 \\
LLMLingua-2 & 32.02 & 32.24 & 24.99 & 27.87  & 17.67  & 52.63  & 30.75  \\
CaM & 32.14 & 32.63 & 24.91 &63.09 & 16.77 & 54.03 & 37.49 \\
\cellcolor{highlight}{\textbf{AsymKV}} & \cellcolor{highlight}{\textbf{39.42}} & \cellcolor{highlight}{\textbf{38.93}} & \cellcolor{highlight}{\textbf{27.30}} & \cellcolor{highlight}{65.66} & \cellcolor{highlight}{\textbf{39.39}} & \cellcolor{highlight}{55.24} & \cellcolor{highlight}{\textbf{43.95}} \\
\midrule
\multicolumn{8}{c}{\textbf{Mistral-7B-Instruct-v0.3}} \\
\midrule
Full Context & 38.74 & 38.29 & 29.04 & 70.70 & 51.00 & 55.06 & 46.40 \\
StreamingLLM & 24.80 & 22.14 & 25.18 & 66.49 & 15.14 & 53.51 & 34.57 \\
LongCache\textbf{ }& 26.05 & 22.31 & 25.44 & 66.21 & 14.93 & 53.43 & 34.80 \\
H$_2$O & 29.66 & 28.22 & 26.32 & {\bf 67.78} & 14.83 & 53.95 & 37.09 \\
LLMLingua-2 & 28.12 & 28.62 & 25.75 & 45.85 & 16.00 & 48.81 & 32.17 \\
CaM\textbf{ }& 26.15 & 29.06 & 26.81 & 66.16 &20.96 & 53.76 & 37.12 \\
\cellcolor{highlight}{\textbf{AsymKV}} & \cellcolor{highlight}{\textbf{33.71}} & \cellcolor{highlight}{\textbf{32.81}} & \cellcolor{highlight}{\textbf{27.04}} & \cellcolor{highlight}{67.21} & \cellcolor{highlight}{\textbf{34.56}} & \cellcolor{highlight}{\textbf{54.93}} & \cellcolor{highlight}{\textbf{41.33}} \\
\midrule
\multicolumn{8}{c}{\textbf{Qwen2-7B-Instruct}} \\
\midrule
Full Context & 37.07 & 41.77 & 28.27 & 68.61 & 36.25 & 50.67 & 43.81 \\
StreamingLLM & 27.73 & 28.14 & 24.32 & 66.85 & 7.50 & 49.55 & 34.70 \\
LongCache & 27.98 & 28.98 & 24.80 & 66.38 & 9.00 & 48.34 & 34.94 \\
H$_2$O & 29.91 & 28.49 & 25.21 & 68.08 & 12.25 & \textbf{53.63} & 36.68 \\
LLMLingua-2 & 30.04 & 31.71 & 24.63 & 46.32 & 6.50 & 50.09 & 31.96 \\
CaM & 29.14 & 28.59 & 25.87 & 66.33 & 9.25 & 48.88 & 35.38  \\
\cellcolor{highlight}{\textbf{AsymKV}} & \cellcolor{highlight}{\textbf{33.72}} & \cellcolor{highlight}{\textbf{34.46}} & \cellcolor{highlight}{\textbf{26.24}} & \cellcolor{highlight}{\textbf{69.88}} & \cellcolor{highlight}{\textbf{14.25}} & \cellcolor{highlight}{49.33} & \cellcolor{highlight}{\textbf{38.76}} \\
\bottomrule
\end{tabular}
\end{adjustbox}
\label{tab:longbench}
\end{table*}

\subsection{Experimental Setup}  

%We conduct comprehensive evaluations to assess AsymKV's effectiveness in long-context scenarios. 

{\bf Baselines} We compare AsymKV against several categories of approaches: {\it KV cache compression}: H$_2$O~\cite{zhang2023h2o}, {\it KV cache merge}: CaM~\cite{zhangcam} ,{\it prompt compression}: LLMLingua-2.0~\cite{pan-etal-2024-llmlingua}, {\it context segmentation}: StreamingLLM~\cite{xiao2023streamingllm} and LongCache~\cite{liu2024farewell}. %We also use standard full-context inference without compression serves as our performance upper bound.  

{\bf Base Models}  To demonstrate the generality of AsymKV, we evaluate across diverse model architectures: Llama2-7B-chat~\cite{touvron2023llama}, Llama3.1-8B-Instruct~\cite{dubey2024llama}, Mistral-7B-Instruct-v0.3~\cite{jiang2023mistral7b}, and Qwen2-7B-Instruct~\cite{yang2024qwen2technicalreport}.% Notably, AsymKV requires no model fine-tuning, making it immediately applicable to existing LLM deployments. 

%\textbf{Implementation Details} Unless otherwise specified, we set the compression context \texttt{max\_length} to $2048$ tokens and \texttt{chunk} to $512$. Following attention sink~\cite{xiao2023streamingllm}, we always preserve the initial $32$ tokens. All experiments are conducted on a single NVIDIA A100 80GB GPU to ensure consistent and reproducible measurements.

\textbf{Implementation Details} Unless otherwise specified, we set the compression context \texttt{max\_length} to $2048$ tokens and \texttt{chunk\_size} to $512$. All baselines use the same settings for fair comparison. For H$_2$O, we set recent budget to $2048$ and heavy budget to $512$. Following attention sink~\cite{xiao2023streamingllm}, we always preserve the initial $32$ tokens. All experiments are conducted on NVIDIA A100 80GB.

\subsection{Long Context Performance Evaluation}

%We evaluate the performance of stated-CLM in long-context QA tasks to validate its ability to memory contexts. The experiments are conducted using LongBench~\citep{bai2023longbench}, a benchmark specifically designed for long-context scenarios. Given that AsymKV supports unlimited context lengths, we do not truncate the input data in our experiments except for the full context baseline.

We evaluate AsymKV's effectiveness on LongBench~\citep{bai2024longbench}, a comprehensive benchmark for long-context understanding. LongBench contains 16 English tasks from a wide range of categories. %Given AsymKV's ability to handle unlimited context lengths, we process inputs without truncation, except for the full context baseline.  

\textbf{Results} As shown in Table~\ref{tab:longbench}, AsymKV consistently outperforms existing long-context methods across different models and tasks. On LLaMA3.1-8B, AsymKV achieves a 43.95 average score, surpassing H$_2$O (38.89) and other baselines by a significant margin. The improvement is particularly pronounced in challenging tasks like Synthetic reasoning, where AsymKV scores 39.39 compared to H$_2$O's 14.75. For all base models, AsymKV maintains its advantage with average scores, showing substantial improvements over baselines.  

\textbf{Precise Information Retrieval} In Single-Doc and Multi-Doc QA tasks, which require precise information retention, AsymKV consistently outperforms other methods by significant margins (5-10 points). This suggests that our homogeneity-based compression effectively preserves key information needed for accurate question answering. 

%\textbf{Long-Context Reasoning} The strong performance in Synthetic tasks (improvements of up to 25 points over baselines) indicates that AsymKV maintains complex reasoning chains that other compression methods might disrupt. Meanwhile, the relatively smaller gains in Code tasks suggest that structured content with rigid syntax patterns presents a different challenge for compression strategies.  

\textbf{Extreme Long-Context Compression} We evaluate AsymKV on LongBenchV2~\cite{bai2024longbench2}, a benchmark with contexts ranging from 8,000 to 2 million tokens across six task categories (multi-document QA, code comprehension, temporal reasoning, mathematical derivation, cross-lingual understanding, and hierarchical information synthesis). Using Llama3.1-8B-Instruct with cache\_size=8192, Table~\ref{tab:LongBenchV2} shows that AsymKV matches full-context methods in short contexts while significantly outperforming baselines in medium to long contexts (up to millions of tokens), demonstrating its effectiveness in extreme long-context senarios.

\begin{table}[t]
    \small
    \centering
    \caption{Performance on LongBenchV2.}
    \begin{tabular}{lcccccc}
        \hline
        \textbf{Model} & \textbf{Overall} & \textbf{Easy} & \textbf{Hard} & \textbf{Short} & \textbf{Medium} & \textbf{Long} \\
        \hline
        Full Context & 30.02 & 30.73 & 29.58 & 35.00 & 27.91 & 25.93 \\
        StreamingLLM & 27.04 & 27.60 & 26.69 & 32.78 & 23.26 & 25.00 \\
        LongCache & 28.43 & 28.13 & 28.62  & 32.78 & 25.58 & 26.85  \\
        H$_2$O   &  28.23  & 28.12 & 28.29 & 31.67 & 26.98 & 25.00\\
        CaM & 28.23 & 28.64 & 27.97 & 31.67 & 26.98 & 25.00\\
        \cellcolor{highlight}\textbf{AsymKV} & \cellcolor{highlight}\textbf{30.02} & \cellcolor{highlight}\textbf{30.23} & \cellcolor{highlight}\textbf{29.90} & \cellcolor{highlight}\textbf{32.78} & \cellcolor{highlight}\textbf{27.44} & \cellcolor{highlight}\textbf{28.85} \\
        \hline
    \end{tabular}
    \label{tab:LongBenchV2}
\end{table}

{\bf Regularization Effect of AsymKV.} 
AsymKV often outperforms the full-context model across various tasks, suggesting that it acts as a form of regularization in long-context settings. Due to the inherent limitations of LLMs in handling extended contexts, full KV caches tend to accumulate redundant tokens with low attention scores, diluting focus on relevant information. By selectively merging these low-attention tokens, AsymKV effectively suppresses contextual noise, leading to more focused and efficient inference. A similar regularization phenomenon has also been observed in related KV-cache optimization methods~\cite{zhang2023h2o}.

\subsection{Comprehensive Information Retention} 

KV cache compression methods face multiple challenges in information retention: they must preserve not only early context details but also maintain the ability to capture document-level semantic structure and sequential relationships. %Many existing methods struggle with these requirements, particularly when handling multi-topic documents that demand both local detail preservation and global context understanding.  
To evaluate models' comprehensive information retention capabilities, we conduct experiments using TopicRet~\citep{li2023long} from L-Eval~\citep{an-etal-2024-l}. This benchmark is particularly challenging as it requires models to answer questions about {\it the second or third topic in multi-topic documents}, testing their ability to retain early context information. %, (2) maintain document-level semantic understanding for topic differentiation, and (3) preserve sequential relationships between topics. 

% \begin{table}[ht]  
%     \small
%     \centering  
%     \setlength{\tabcolsep}{4pt}
%     \caption{Performance on early topic retrieval. %Results demonstrate AsymKV's superior ability to retain and access early context information compared to other compression methods.
%     }  
%     \begin{adjustbox}{max width=\linewidth}  
%     \begin{tabular}{lcccccc}  
%     \toprule  
%     \textbf{Model} & \textbf{Qwen2-7B} & \textbf{Llama3.1-8B}& \textbf{Mistral-7B} \\
%     \midrule  
%     Full Context & 47.33 & 80.00 & 42.67 \\
%     StreamingLLM & 0.00 & 0.00 & 0.00 \\
%     LongCache & 0.00 & 0.00 & 8.67 \\
%     CaM & 12.67 & 24.67 & 15.33 \\
%     H$_2$O & 21.33 & 63.33 & 38.67 \\
%     LLMLingua2 & 0.00 & 0.00 & 0.00 \\
%     \cellcolor{highlight}\textbf{AsymKV} & \cellcolor{highlight}\textbf{36.00} & \cellcolor{highlight}\textbf{75.33} & \cellcolor{highlight}\textbf{40.67} \\
%     \bottomrule  
%     \end{tabular}  
%     \end{adjustbox}  
%     \label{tab:topicret}  
% \end{table}  

\begin{figure}[t]
    \centering
    \begin{minipage}{0.6\textwidth}
        \small
        \centering  
        \setlength{\tabcolsep}{4pt}
        \captionof{table}{Performance on early topic retrieval. %Results demonstrate AsymKV's superior ability to retain and access early context information compared to other compression methods.
        }  
        \begin{adjustbox}{max width=\linewidth}  
        \begin{tabular}{lcccccc}  
        \toprule  
        \textbf{Model} & \textbf{Qwen2-7B} & \textbf{Llama3.1-8B}& \textbf{Mistral-7B} \\
        \midrule  
        Full Context & 47.33 & 80.00 & 42.67 \\
        StreamingLLM & 0.00 & 0.00 & 0.00 \\
        LongCache & 0.00 & 0.00 & 8.67 \\
        CaM & 12.67 & 24.67 & 15.33 \\
        H$_2$O & 21.33 & 63.33 & 38.67 \\
        LLMLingua2 & 0.00 & 0.00 & 0.00 \\
        \cellcolor{highlight}\textbf{AsymKV} & \cellcolor{highlight}\textbf{36.00} & \cellcolor{highlight}\textbf{75.33} & \cellcolor{highlight}\textbf{40.67} \\
        \bottomrule  
        \end{tabular}  
        \end{adjustbox}  
        \label{tab:topicret}  
    \end{minipage}
    \hfill
    \begin{minipage}{0.35\textwidth}
        \centering
        \includegraphics[width=0.99\linewidth]{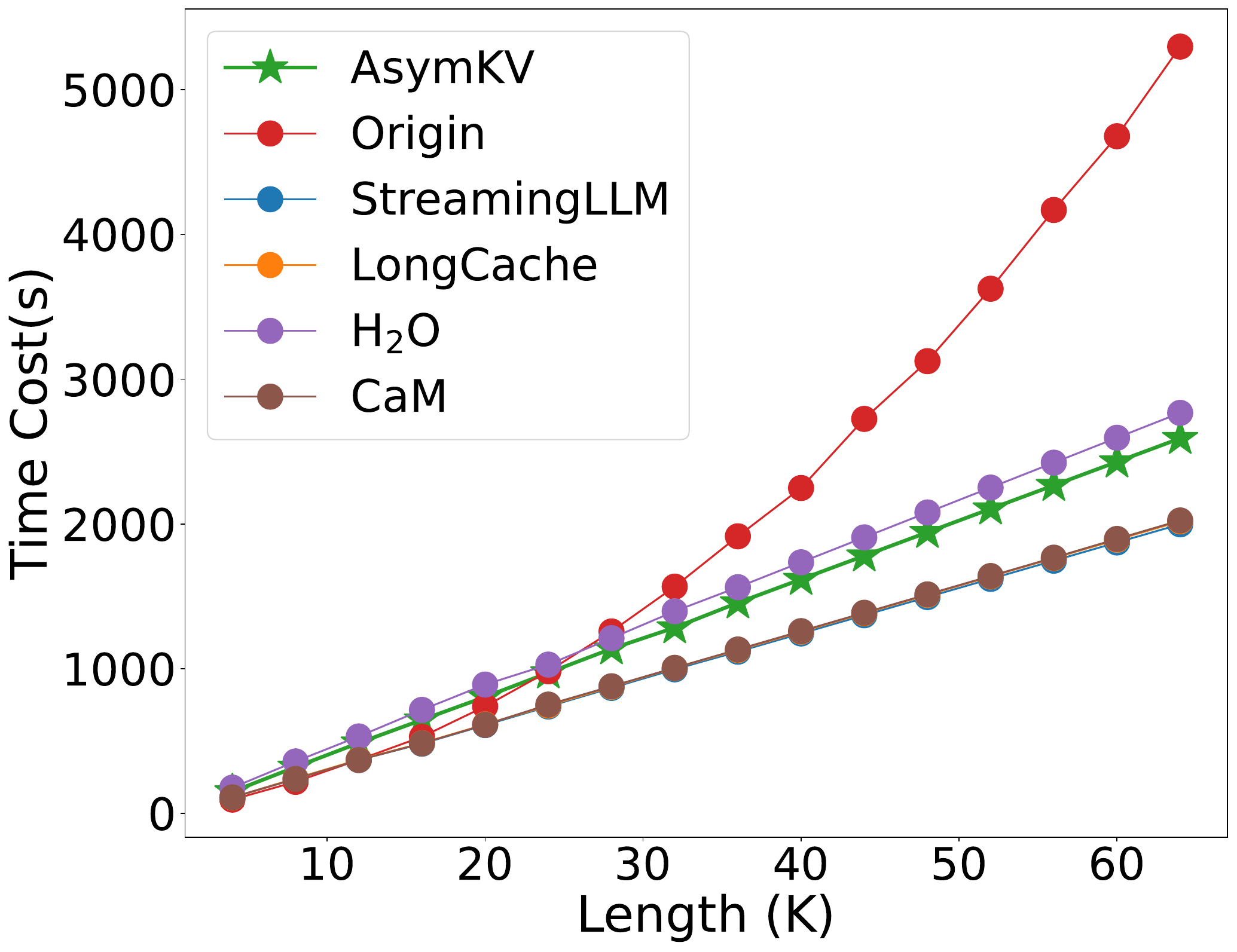}
        \caption{Inference efficiency.}
        \label{fig:inference_efficiency_comparison}
    \end{minipage}
\end{figure}

The results in Table~\ref{tab:topicret} reveal several key findings. First, context segmentation methods (StreamingLLM, LongCache) and prompt compression approaches (LLMLingua2) completely fail at this task, scoring zero or near-zero across all models. This dramatic performance drop confirms our hypothesis that discarding or imprecisely compressing early tokens severely impairs models' ability to access historical information. Although methods like CaM and H$_2$O show some capability in retaining early information, their performance significantly lags behind full-context processing. In contrast, AsymKV demonstrates remarkable effectiveness in preserving early context information, achieving scores close to full-context processing (75.33 vs 80.00 on LLaMA3.1-8B). %This strong performance can be attributed to our asymmetric compression strategy. %: our cardinality-aware value representation ensures that all values are retained. %These results validate that AsymKV can maintain precise token-level information even after substantial compression, enabling accurate retrieval of early context details when needed.

\subsection{Compression Rate Analysis}
To systematically evaluate AsymKV's context compression capabilities, we analyze its performance across different compression rates on the long-context HotpotQA~\citep{yang2018hotpotqa} task from LongBench. Here, the compression rate is defined as the ratio between the compressed and original token counts.

\begin{figure}[t]
    \centering
    \begin{subfigure}[b]{0.245\columnwidth}
        \includegraphics[width=\textwidth]{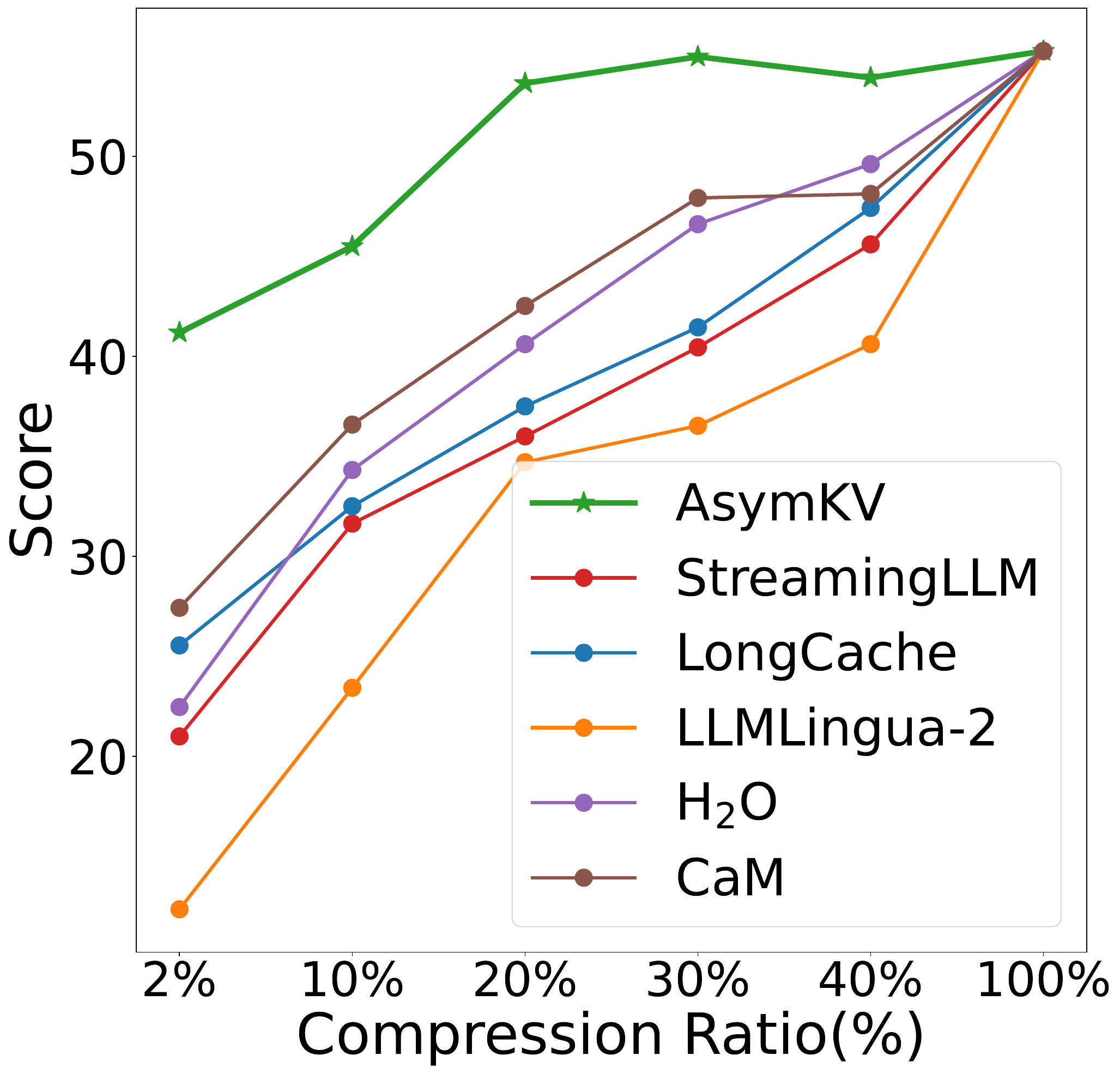}
        \caption{Llama-3.1-8B-Instruct}
        \label{fig:llama3.1-8b}
    \end{subfigure}
    \hfill
    \begin{subfigure}[b]{0.245\columnwidth}
        \includegraphics[width=\textwidth]{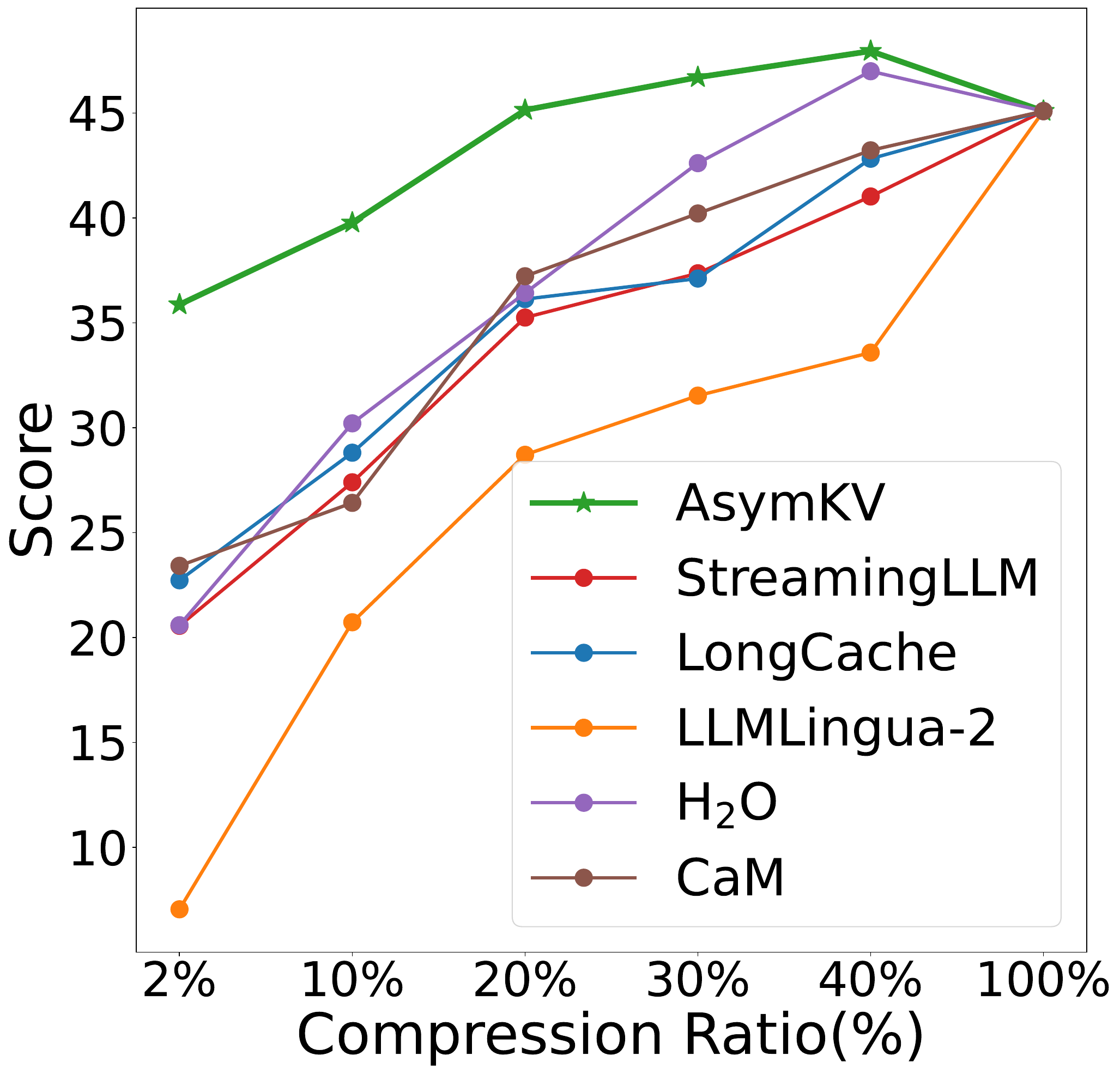}
        \caption{Mistral-7B-Inst.-v0.3}
        \label{fig:mistral}
    \end{subfigure}
    \begin{subfigure}[b]{0.245\columnwidth}
        \includegraphics[width=\textwidth]{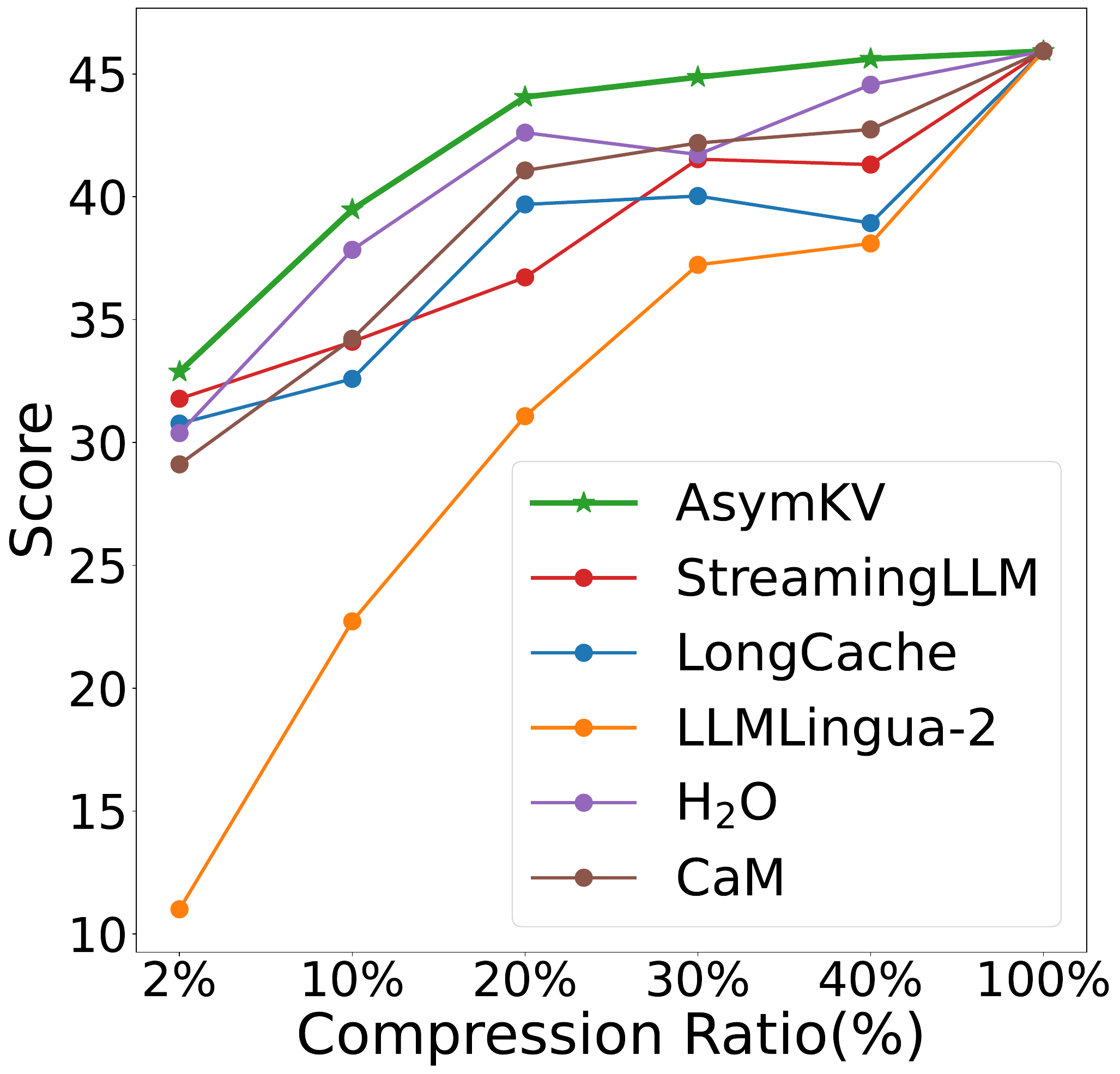}
        \caption{Qwen2-7b-Instruct}
        \label{fig:Qwen2}
    \end{subfigure}
    \hfill
    \begin{subfigure}[b]{0.245\columnwidth}
        \includegraphics[width=\textwidth]{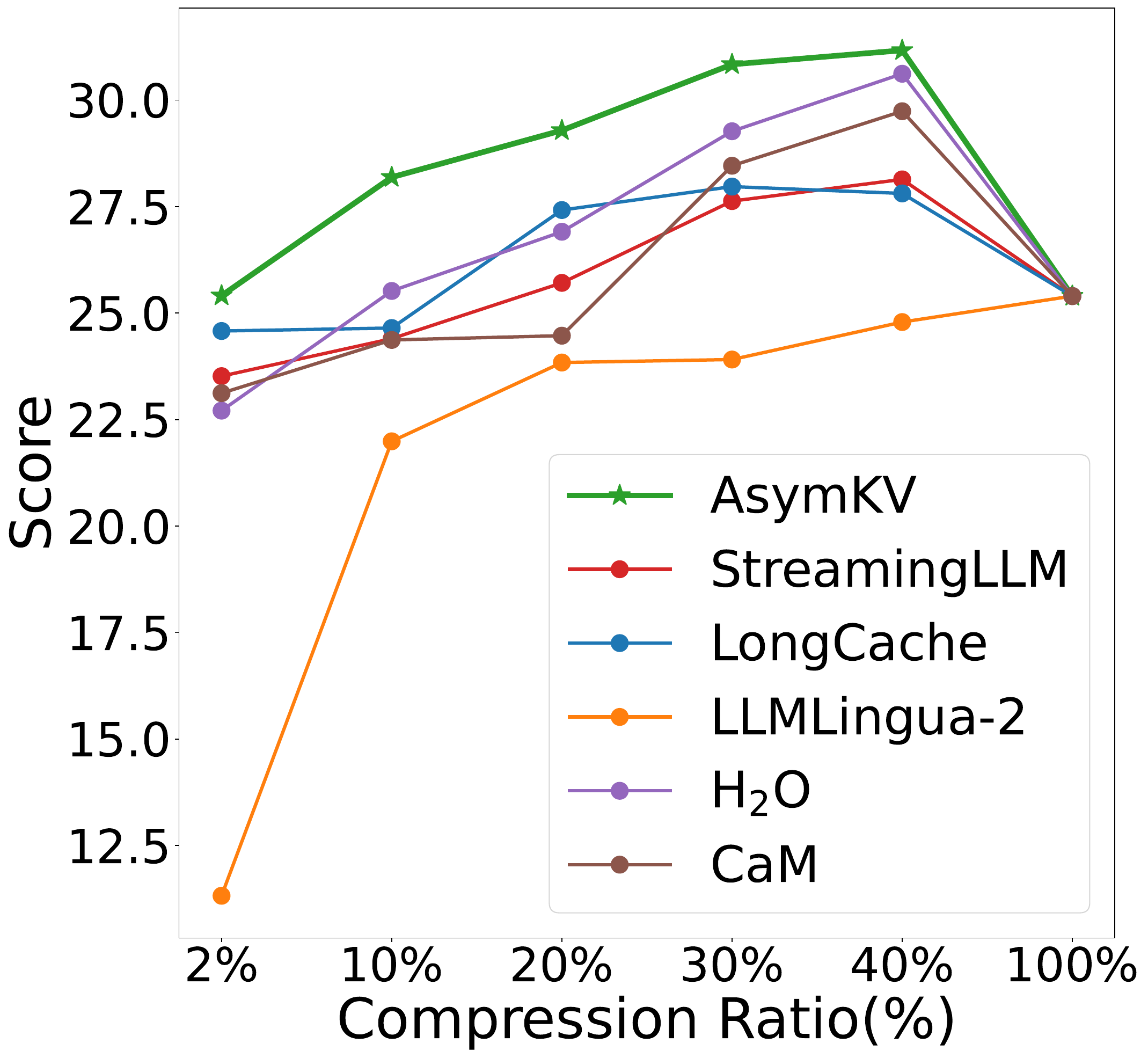}
        \caption{Llama-2-7b-chat-4k}
        \label{fig:llama2-7b}
    \end{subfigure}
    \caption{Effect of different compression ratios. %Even when compressed to only 10\% of the original context length, the model can still maintain performance close to that of the full context.
    }
    \label{fig:compression_rate}
\end{figure}

As shown in Fig.~\ref{fig:compression_rate}, AsymKV demonstrates superior compression capabilities across all compression rates. Most notably, with only 20\% of the original context length, AsymKV achieves performance comparable to full-context processing, significantly outperforming all baseline methods. This robust performance highlights AsymKV's effectiveness in preserving crucial contextual information even under aggressive compression.

%Interestingly, at certain compression rates (e.g., 40\% of original length), AsymKV's performance actually exceeds that of full-context processing (100\%). This counter-intuitive result suggests that AsymKV not only effectively preserves essential information but also acts as an intelligent filter, potentially reducing noise and helping the model focus on the most relevant context tokens.

% \begin{figure}[t]
%     \centering  
%     %\begin{subfigure}[b]{0.43\textwidth}  
%         \includegraphics[width=0.3\textwidth]{figs/Mistral-7B-Instruct-v0.3-time.pdf}  
%         %\caption{Mistral-7B-Instruct-v0.3}  
%         \label{fig:mistral-e}  
%     %\end{subfigure}  
%     \caption{Inference time across different methods. %While context segmentation approaches achieve highest efficiency due to minimal KV cache operations, AsymKV demonstrates superior efficiency among compression-based methods while maintaining high performance.
%     }  
%     \label{fig:inference_efficiency_comparison}  
% \end{figure}  

\begin{figure}[t]
    \centering
    \begin{minipage}{0.4\textwidth}
        \small
        \centering
        \captionof{table}{Peak GPU Memory (MB)}
        \begin{adjustbox}{max width=\linewidth}
        \begin{tabular}{lcc}
        \toprule
        \textbf{Method} & \textbf{L.2-7B} & \textbf{L.2-13B} \\
        \midrule
        StreamingLLM & 19,592 & 39,911\\
        LongCache & 24,968 & 49,236\\
        H$_2$O & 22,479 & 47,310\\
        CaM & 22,548 & 47,476\\
        AsymKV & 24,923 & 48,671\\
        \bottomrule
        \end{tabular}
        \end{adjustbox}
        \label{tab:memory_consumption}
    \end{minipage}
    \hfill
    \begin{minipage}{0.56\textwidth}
        \small
        \centering  
        \setlength{\tabcolsep}{4pt}
        \caption{Ablation study on different merge strategies. %For value merging, ``Key-identical'' indicates using the same strategy as keys, while ``Asymmetric'' uses our proposed cardinality-norm method.
        }  
        \begin{adjustbox}{max width=\linewidth}  
        \begin{tabular}{lcc}  
        \toprule  
        & \multicolumn{2}{c}{\textbf{Value Merge Strategy}} \\
        \cmidrule(lr){2-3}  
        \textbf{Key Merge Strategy} & \textbf{Same as Key} & \textbf{Asymmetric} \\
        \midrule  
        Mean Merge & 18.93& 21.86\\
        Weighted by Cardinality & 20.90 & 23.91\\
        Weighted by Attention & 14.99& 23.83\\
        \cellcolor{highlight}\textbf{Optimal $\mathbf{k}^*$ (Eq.~\eqref{eq:optimal_k})} & \cellcolor{highlight}21.21& \cellcolor{highlight}\textbf{24.43}\\
        \bottomrule  
        \end{tabular}  
        \end{adjustbox}  
        \label{tab:merge_ablation}  
    \end{minipage}
\end{figure}

\subsection{Inference Efficiency}  
\label{sec:exp:efficiency}

We evaluated the computational efficiency of different approaches during text generation. To do this, we had the models generate text using greedy sampling on Mistral-7B-Instruct-v0.3 and measured the time required to generate different numbers of tokens. 

{\bf Inference Speed} Fig.~\ref{fig:inference_efficiency_comparison} reveals that among the evaluated methods, context segmentation approaches (StreamingLLM and LongCache) achieve the highest computational efficiency due to their minimal KV cache operations. However, this efficiency comes at the cost of performance. AsymKV strikes a better balance, achieving the highest efficiency among compression-based methods.

% \begin{table}[t]  
% \small
% \centering  
% \caption{Ablation study on different merge strategies for keys and values. For value merging, ``Key-identical'' indicates using the same strategy as keys, while ``Cardinality-norm'' uses our proposed normalization method. Results are scores on the multi-hop QA task MuSiQue~\cite{trivedi2022musique} from LongBench.}  
% \begin{adjustbox}{max width=\linewidth}  
% \begin{tabular}{lcc}  
% \toprule  
% & \multicolumn{2}{c}{\textbf{Value Merge Strategy}} \\
% \cmidrule(lr){2-3}  
% \textbf{Key Merge Strategy} & \textbf{Key-identical} & \textbf{Cardinality-norm} \\
% \midrule  
% Mean Merge & 18.93& 21.86\\
% Weighted by Cardinality & 20.90 & 23.91\\
% Weighted by Attention Score & 14.99& 23.83\\
% \cellcolor{highlight}\textbf{Optimal $\mathbf{k}^*$ (Eq.~\eqref{eq:optimal_k})} & \cellcolor{highlight}21.21& \cellcolor{highlight}\textbf{24.43}\\
% \bottomrule  
% \end{tabular}  
% \end{adjustbox}  
% \label{tab:merge_ablation}  
% \end{table}  

%\subsection{Memory Consumption}  
%\label{sec:exp:mem}
{\bf Memory Consumption } 
We measure peak GPU memory usage on LLaMA3.1-8B-Instruct with a cache size of 2048 and chunk size of 128. 
As shown in Table~\ref{tab:memory_consumption}, AsymKV's memory consumption is comparable to other baselines. This validates that both generation and compression costs are practical and scalable.

% \begin{table}[t]  
% \centering  
% \caption{Peak GPU Memory Consumption (MB) on LLaMA3.1-8B-Instruct}  
% \begin{adjustbox}{max width=\linewidth}  
% \begin{tabular}{lcccccc}  
% \toprule  
% \textbf{Method} & StreamingLLM & LongCache & H₂O & D₂O & CaM & AsymKV \\  
% \midrule  
% \textbf{Peak Memory} & 20,458 & 22,319 & 21,458 & 22,743 & 21,072 & 23,930 \\  
% \bottomrule  
% \end{tabular}  
% \end{adjustbox}  
% \label{tab:memory_consumption}  
% \end{table}  

\subsection{Ablations on Merge Strategies}  

We conduct an ablation study to compare different strategies for merging keys and values during compression. For key merging, we compare four approaches: simple mean pooling, cardinality-weighted averaging, attention score-weighted averaging, and our optimal strategy derived from Eq.~\eqref{eq:optimal_k}. For each key merge strategy, we experiment with two value merge strategies: either using the identical strategy as keys, or using our proposed asymmetric cardinality-normalized method.  

Results in Table~\ref{tab:merge_ablation} are scores on the multi-hop QA task MuSiQue~\cite{trivedi2022musique} from LongBench. The results demonstrate two key findings. First, our theoretically-derived optimal key merge strategy consistently outperforms other approaches. 
%When combined with cardinality-normalized value merging, it achieves the best performance of 24.43, significantly higher than simpler alternatives like mean merge (21.86) or attention score-based merge (23.83). 
This empirically validates the foundation of our optimal merging strategy presented in \S~\ref{sec:method:key}.  
Second, the results show that using distinct strategies for keys and values is beneficial. %Across all key merge methods, applying our cardinality-normalized approach to values consistently yields better results than using identical strategies.  

%% file: conclu.tex
\section{Conclusion}  
\label{sec:conclu}

In this paper, through extensive empirical analysis, we reveal a fundamental yet previously overlooked pattern: local KV cache Asymmetry. This property motivates our key technical innovation—a training-free merging framework that combines homogeneity-based key merging with mathematically proven lossless value representation. We present AsymKV, a novel approach to address the computational challenges of long-context modeling in LLMs. Our comprehensive experiments demonstrate that AsymKV outperforms existing long-context methods across various tasks and base models. 

\paragraph{Limitations.} Further applications of AsymKV need to consider compatibility with methods like FlashAttention and vLLM. We view this as an engineering problem and are actively working to address it.

\newpage

\paragraph{Acknowledgments and Disclosure of Funding}
This paper was supported by the Shanghai Natural Science Foundation (25ZR1402137).

%We believe our discovery of the asymmetric distributional properties between keys and values provides valuable insights for future research in long-context modeling. This fundamental understanding of attention mechanisms could inspire new approaches to KV cache management and compression, potentially leading to more efficient and effective solutions.

%% file: analysis.tex
\section{Analysis: Key-Value Asymmetry in Attentions}  
\label{sec:analysis}  

\begin{figure*}[!htbp]
    \centering  
    \begin{subfigure}[b]{0.24\textwidth}  
        \centering  
        \includegraphics[width=\textwidth]{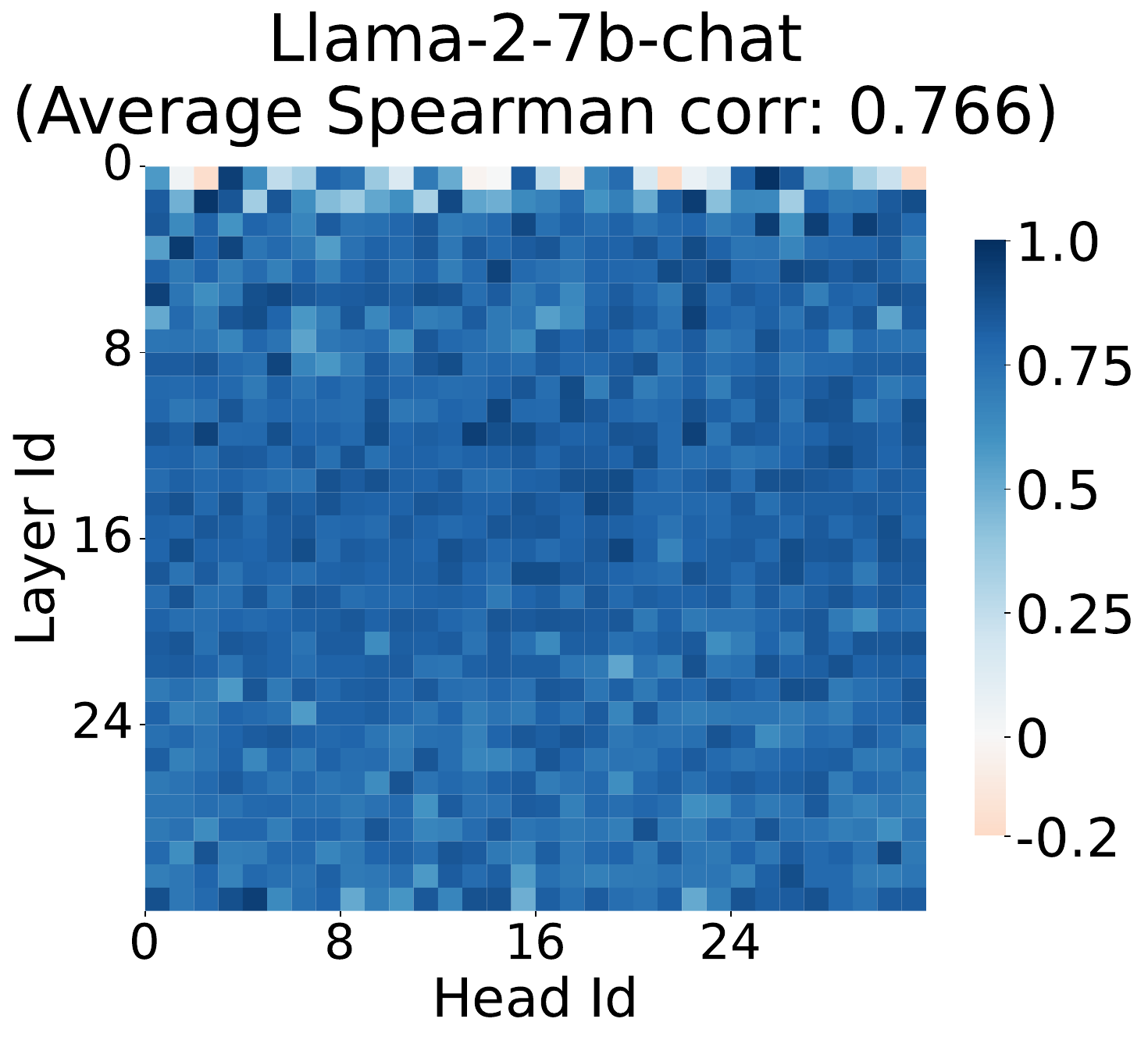}  
        \caption{Key similarity correlation heatmap on QASPER}  
        \label{fig:qasper_key}  
    \end{subfigure}  
    \hfill  
    \begin{subfigure}[b]{0.24\textwidth}  
        \centering  
        \includegraphics[width=\textwidth]{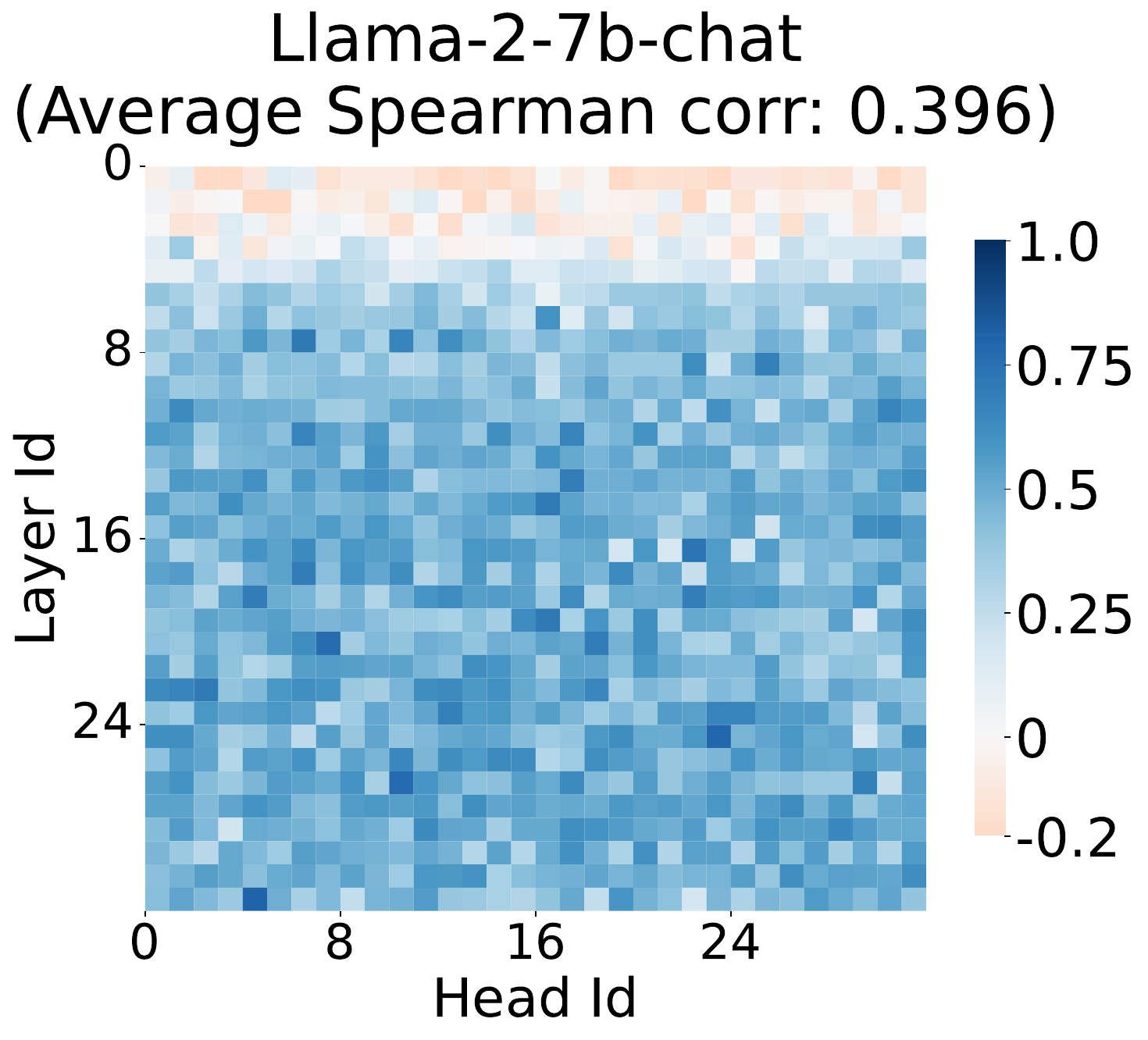}  
        \caption{Value similarity correlation heatmap on QASPER}  
        \label{fig:qasper_value}  
    \end{subfigure}  
    \hfill  
    \begin{subfigure}[b]{0.24\textwidth}  
        \centering  
        \includegraphics[width=\textwidth]{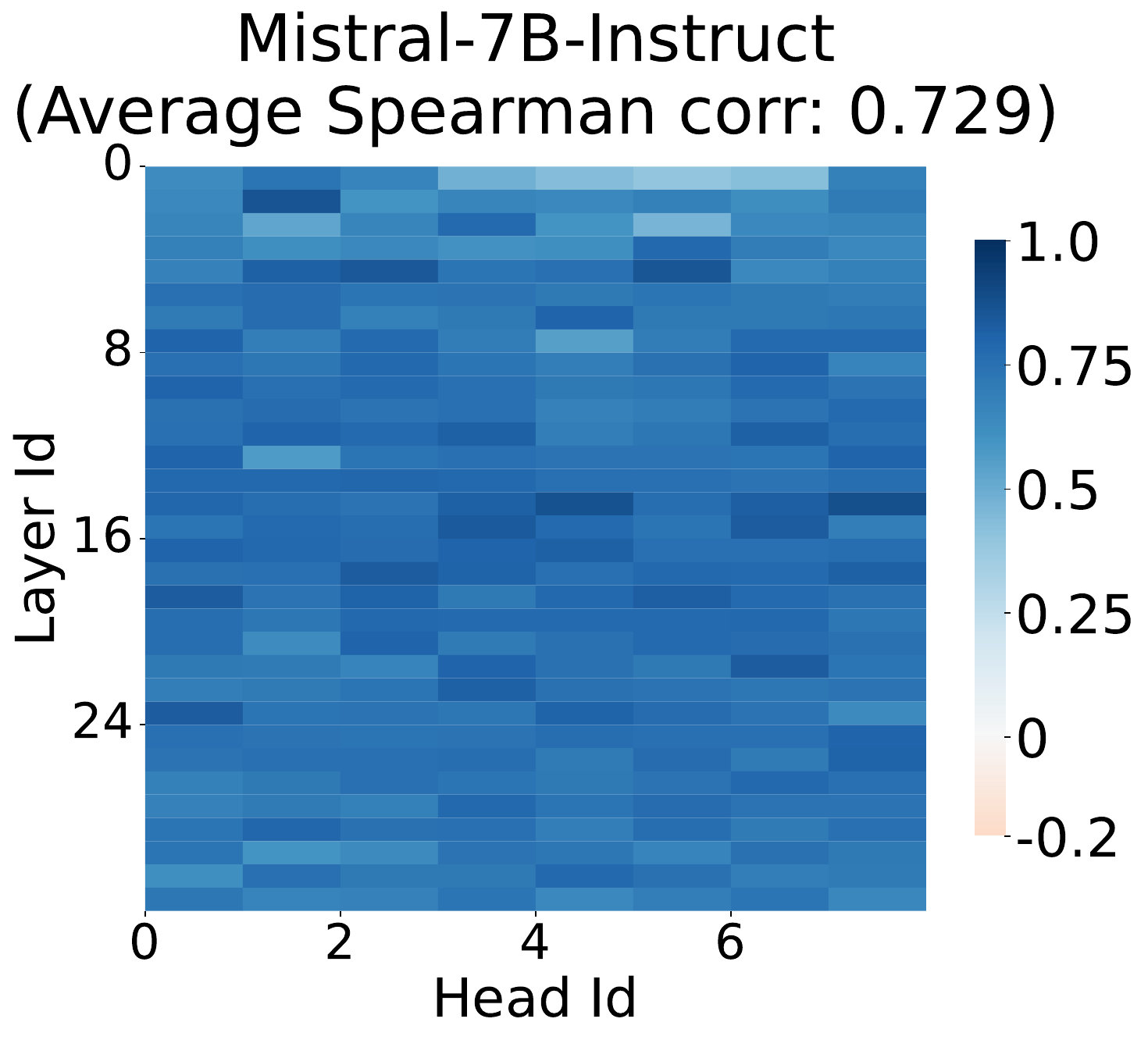}  
        \caption{Key similarity correlation heatmap}  
        \label{fig:multifieldqa_key}  
    \end{subfigure}  
    \hfill  
    \begin{subfigure}[b]{0.24\textwidth}  
        \centering  
        \includegraphics[width=\textwidth]{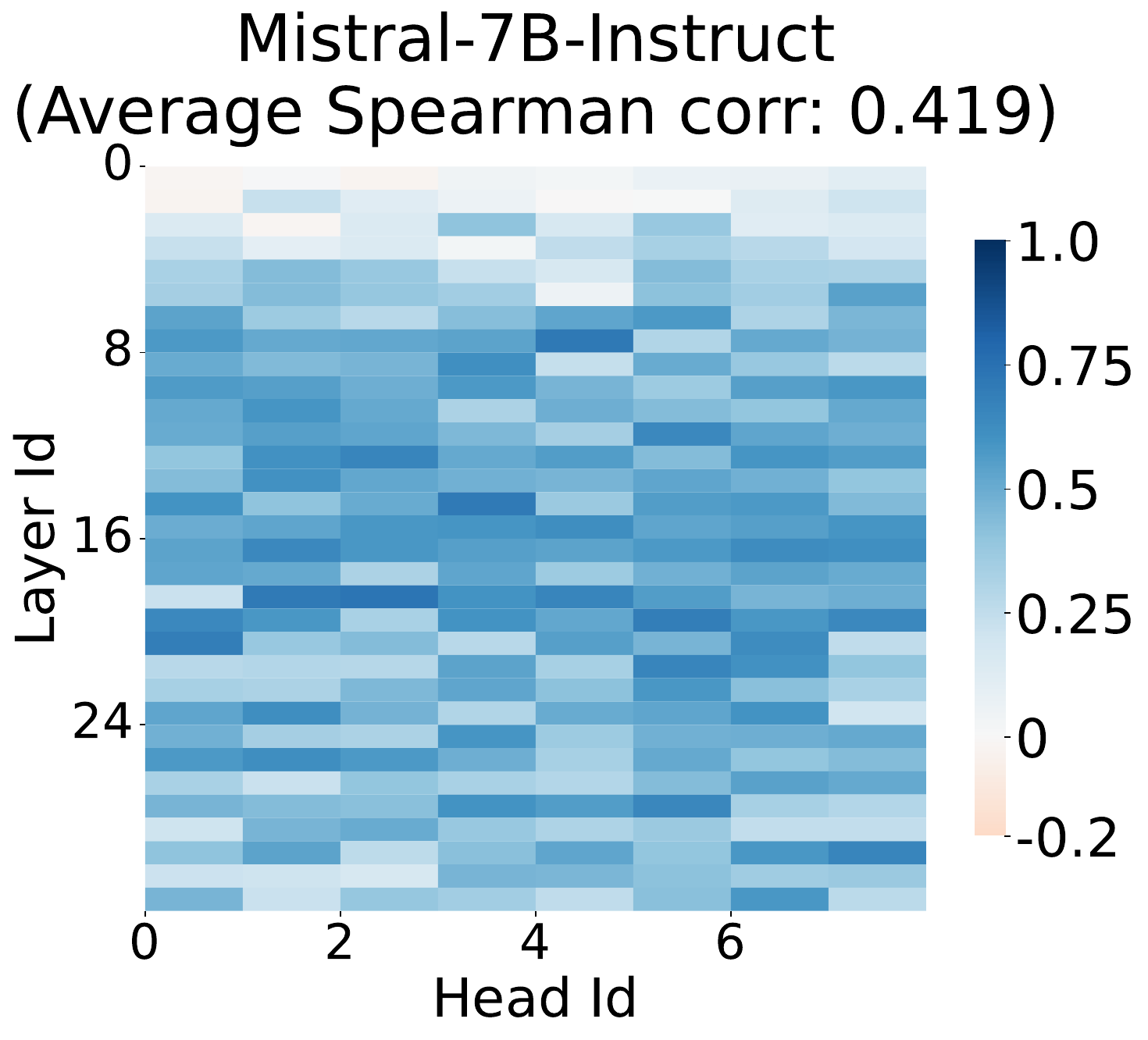}  
        \caption{Value similarity correlation heatmap}   
        \label{fig:multifieldqa_value}  
    \end{subfigure}  
    \caption{Contrasting distributions of local key distributions versus local value distributions across different datasets and model architectures. Heatmaps show Spearman correlation coefficients between adjacent tokens across layers (y-axis) and attention heads (x-axis). The consistent strong positive correlations for local keys (a,c) and weak/negative correlations for local values (b,d) suggesting these are universal properties of LLM KV caches.}  
    \label{fig:distribution}  
\end{figure*}  

%Initial experiments with Llama2-7b-chat on ShareGPT data (Fig.~\ref{fig:all-heatmaps}) revealed a striking asymmetry between local key and value distributions. To validate the universality of these patterns, we conducted a comprehensive analysis across diverse settings: (1) different data distributions, including academic papers (QASPER~\cite{dasigi2021dataset}) and multi-domain questions (MultiFieldQA(en)~\cite{bai2023longbench}) and (2) various model architectures, specifically Mistral-7B-Instruct-v0.3 and Qwen2-7B-Instruct~\cite{yang2024qwen2technicalreport}.  The results for QASPER and Mistral-7B-Instruct-v0.3 are shown in Fig.~\ref{fig:distribution}, with additional results for MultiFieldQA and Qwen2-7B-Instruct presented in Appendix~\ref{sec:append:more_distributions}. 

Initial experiments with Llama2-7b-chat on ShareGPT~\cite{vicuna2023} data (Fig.~\ref{fig:all-heatmaps}) revealed a striking asymmetry between local key and value distributions. To obtain a more direct observation, we analyze the Spearman correlation coefficient of $\text{sim}(\text{key}_i,\text{key}_j)$ and $\text{sim}(\text{val}_i,\text{val}_j)$ for different $i,j$. To validate the universality of these patterns, we conducted a comprehensive analysis across diverse settings: (1) different data distributions, including academic papers (QASPER~\cite{dasigi2021dataset}) and multi-domain questions (MultiFieldQA(en)~\cite{bai2024longbench}) and (2) various model architectures, specifically Mistral-7B-Instruct-v0.3 and Qwen2-7B-Instruct~\cite{yang2024qwen2technicalreport}). The results for QASPER and Mistral-7B-Instruct-v0.3 are shown in Fig.~\ref{fig:distribution}, with additional results for MultiFieldQA and Qwen2-7B-Instruct presented in Appendix~\ref{sec:append:more_distributions}.

\textbf{Key Homogeneity}: Adjacent keys exhibit consistently strong positive correlations across all layers and attention heads (average correlation coefficient $> 0.7$), indicating robust encoding of local semantic relationships in key representations.

\textbf{Value Heterogeneity}: In stark contrast, adjacent values show significantly lower (average correlation coefficient $< 0.4$) or even negative correlations, suggesting that value vectors encode distinct and complementary aspects of token information. This heterogeneity appears essential for maintaining the model's representational capacity.

%These universal patterns have significant implications for KV cache compression. The homogeneity in keys provides strong empirical support for local merging strategies, while the heterogeneity in values indicates the need for a more nuanced approach to value compression. This fundamental asymmetry between keys and values challenges the conventional practice of applying identical compression strategies to both components~\cite{wan2024d2o,wang2024model,zhangcam}.  

\section{Distribution of Local Keys and Values in More Models and Datasets}  
\label{sec:append:more_distributions}  

\begin{figure}[t]  
    \centering  
    \begin{subfigure}[b]{0.24\textwidth}  
        \includegraphics[width=\textwidth]{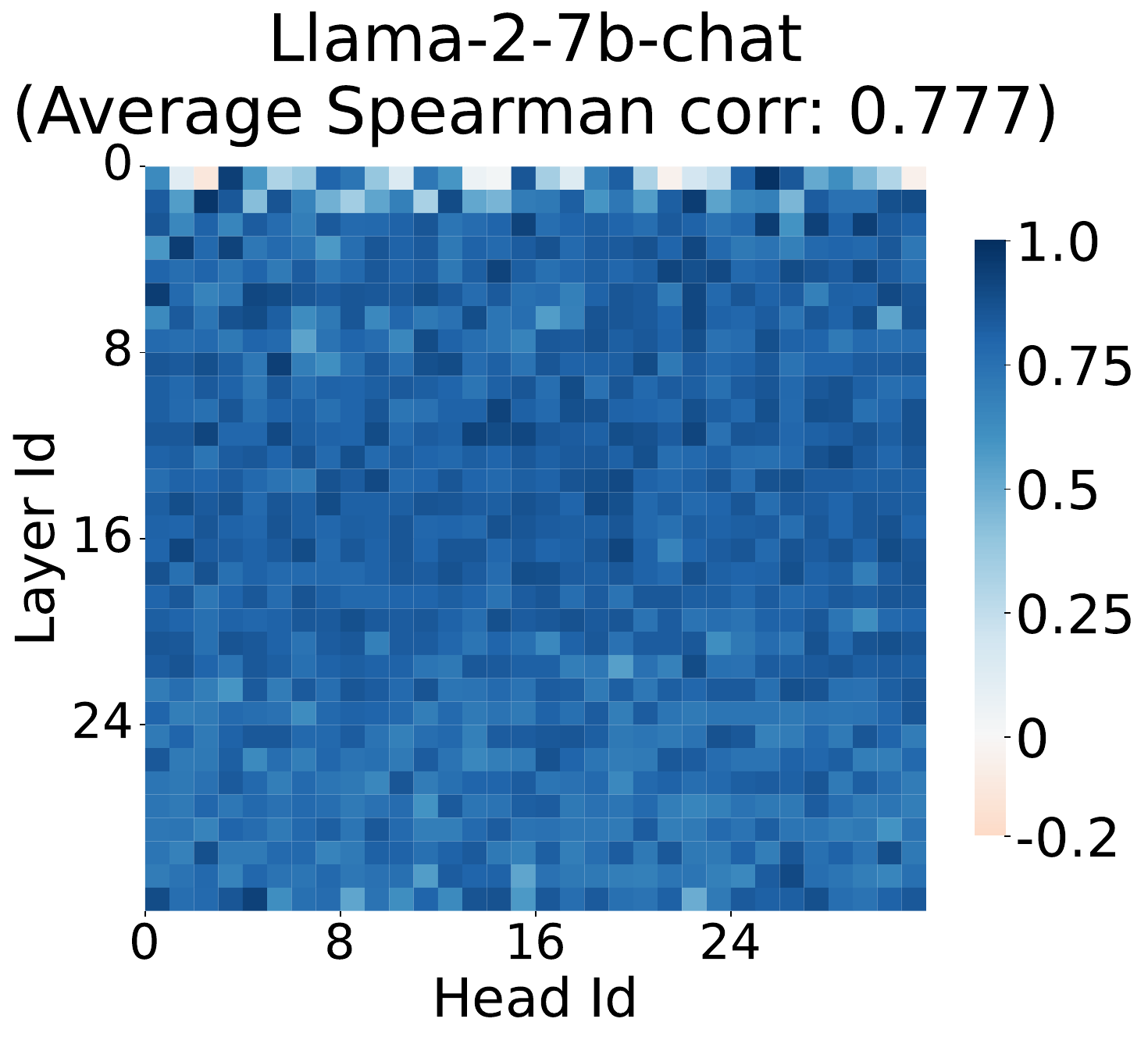}  
        \caption{Key correlation on MultiFieldqa(en)}  
        \label{fig:append:llama_key}  
    \end{subfigure}  
    \hfill  
    \begin{subfigure}[b]{0.24\textwidth}  
        \includegraphics[width=\textwidth]{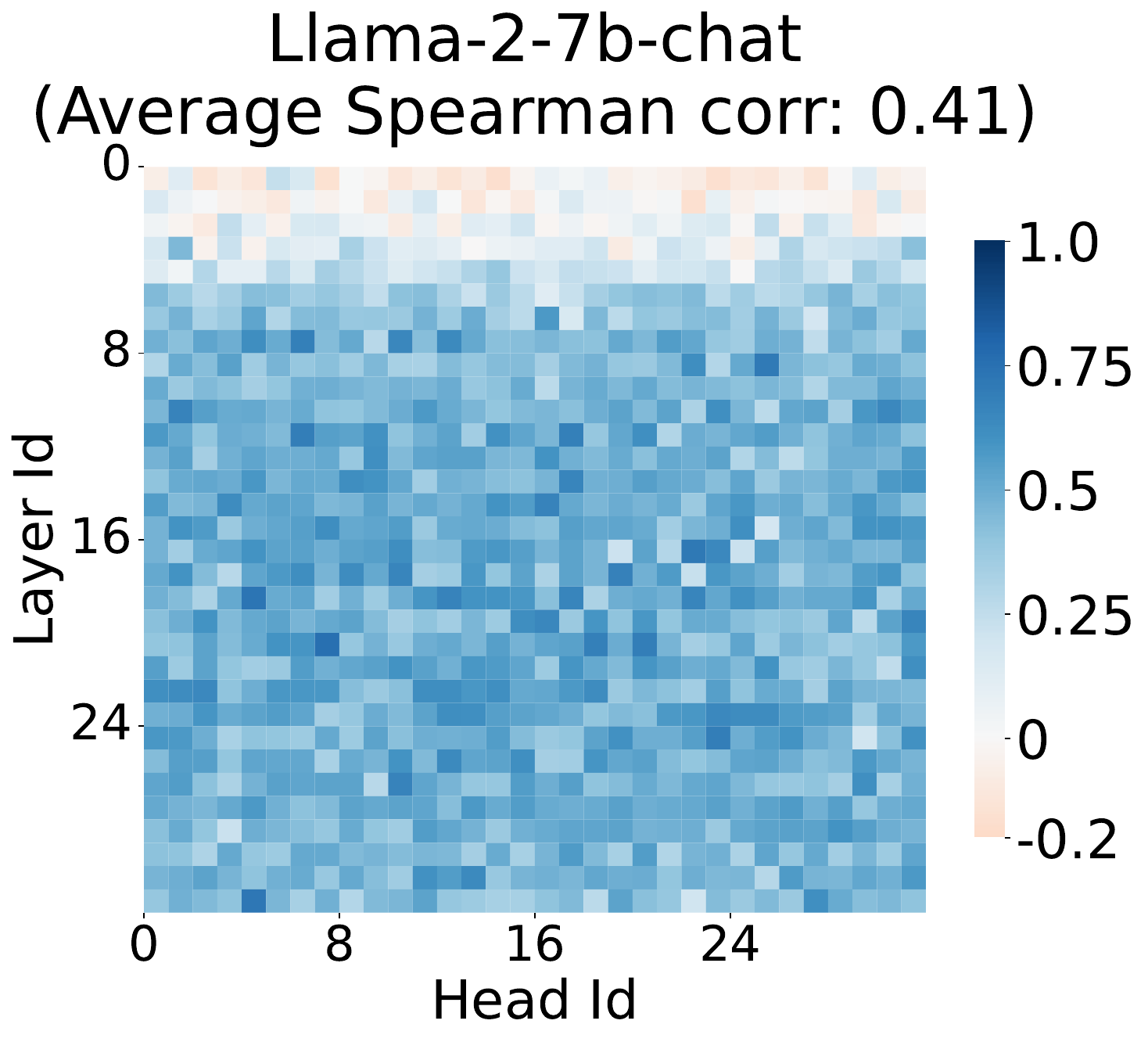}  
        \caption{Value correlation on MultiFieldqa(en)}  
        \label{fig:append:llama_value}  
    \end{subfigure}  
    \hfill  
    \begin{subfigure}[b]{0.24\textwidth}  
        \includegraphics[width=\textwidth]{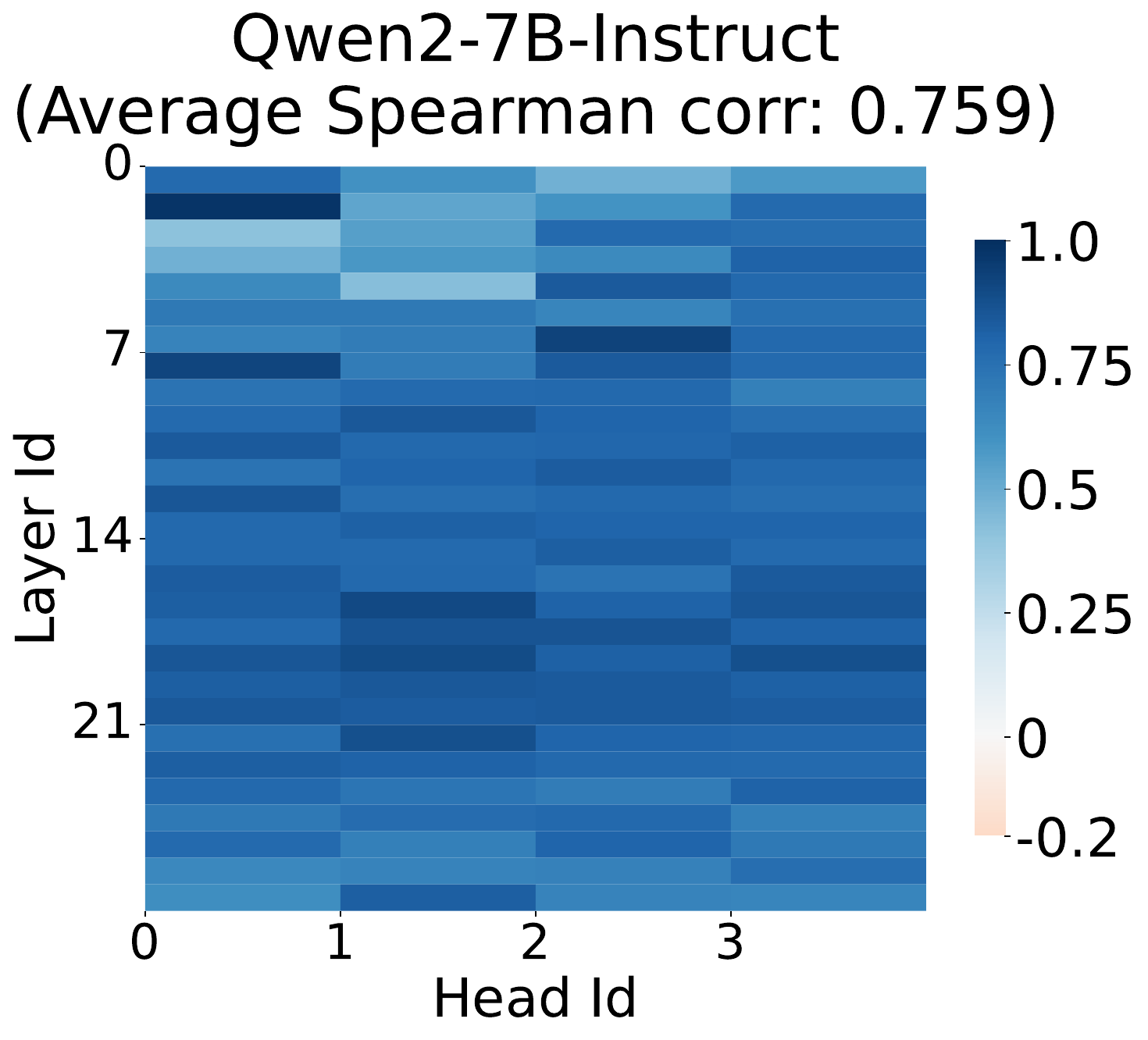}  
        \caption{Key correlation on Qwen2-7B}  
        \label{fig:append:qwen_key}  
    \end{subfigure}  
    \hfill  
    \begin{subfigure}[b]{0.24\textwidth}  
        \includegraphics[width=\textwidth]{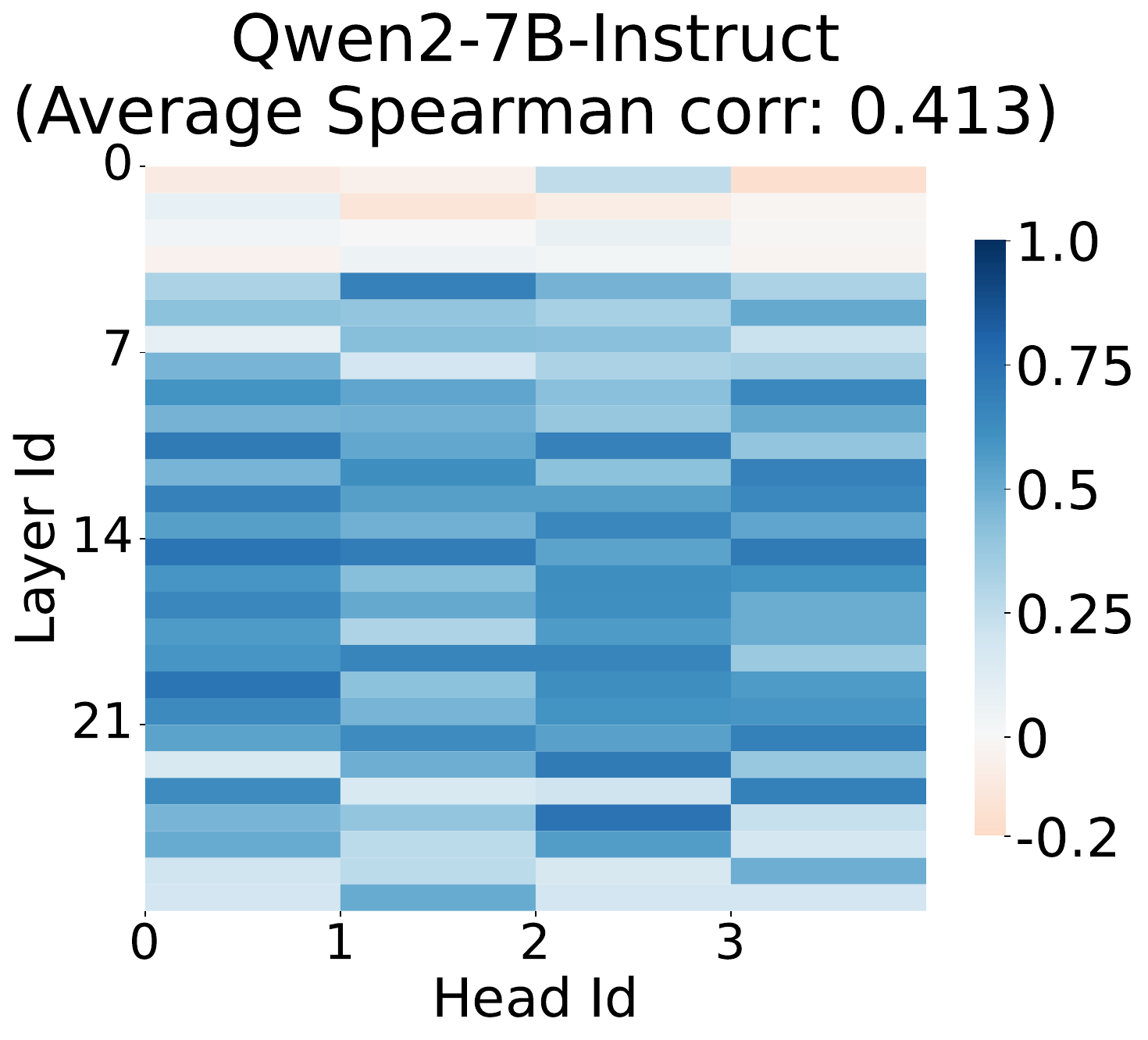}  
        \caption{Value correlation on Qwen2-7B}  
        \label{fig:append:qwen_value}  
    \end{subfigure}  
    \caption{Similarity heatmaps of local keys and values across different models and datasets (LLaMA2-7B-chat on MultiFieldqa(en) and Qwen2-7B-Instruct on ShareGPT). The diagonal blocks in key heatmaps (a, c) indicate strong local homogeneity, while the more scattered patterns in value heatmaps (b, d) demonstrate heterogeneity.}  
    \label{fig:append:more_distributions}  
\end{figure}

To validate that our observations about the asymmetric properties of keys and values are general across different models and datasets, we conduct additional experiments on the Multifieldqa(en) dataset and the Qwen2-7B-Instruct model. The results are shown in Figure~\ref{fig:append:more_distributions}.  

As shown in Figure~\ref{fig:append:more_distributions}, the key similarity heatmaps (Figure~\ref{fig:append:llama_key}, \ref{fig:append:qwen_key}) consistently exhibit strong diagonal block patterns, indicating high local homogeneity. In contrast, the value similarity heatmaps (Figure~\ref{fig:append:llama_value}, \ref{fig:append:qwen_value}) show heterogeneous distributions. These results confirm that the asymmetric properties we observed are inherent characteristics of transformer attention mechanisms rather than artifacts of specific models or datasets.

%% file: append.tex
\section{Solving for the Optimal Key Vector}  
\label{sec:append:solving_for_the_optimal_embedding_vector}

Continue from \eqref{eq:method:optimal_compress:first_step:loss_expansion}:  

\textbf{Gradient Term Expansion:}  
\begin{equation}  
\begin{aligned}  
\nabla \mathcal{L}(\mathbf{k}_m, \mathbf{k}_{m+1})^\top  
\begin{bmatrix}  
\mathbf{k} - \mathbf{k}_m \\
\mathbf{k} - \mathbf{k}_{m+1}  
\end{bmatrix}  
= \mathbf{g}_m^\top (\mathbf{k} - \mathbf{k}_m) + \mathbf{g}_{m+1}^\top (\mathbf{k} - \mathbf{k}_{m+1})  
\end{aligned}  
\end{equation}  

\textbf{Quadratic Term Expansion:}  
\begin{equation}  
\begin{aligned}  
\frac{1}{2}  
\begin{bmatrix}  
\mathbf{k} - \mathbf{k}_m \\
\mathbf{k} - \mathbf{k}_{m+1}  
\end{bmatrix}^\top  
\mathbf{H}  
\begin{bmatrix}  
\mathbf{k} - \mathbf{k}_m \\
\mathbf{k} - \mathbf{k}_{m+1}  
\end{bmatrix}  
= &\frac{1}{2} (\mathbf{k} - \mathbf{k}_m)^\top \mathbf{H}^{11} (\mathbf{k} - \mathbf{k}_m) 
+ (\mathbf{k} - \mathbf{k}_m)^\top \mathbf{H}^{12} (\mathbf{k} - \mathbf{k}_{m+1}) \\
&+ \frac{1}{2} (\mathbf{k} - \mathbf{k}_{m+1})^\top \mathbf{H}^{22} (\mathbf{k} - \mathbf{k}_{m+1})  
\end{aligned}  
\end{equation}  

\textbf{Constructing the Total Objective Function}  

Adding the above terms, the objective function with respect to $\mathbf{k}$ is:  
\begin{equation}  
\begin{aligned}  
\mathcal{L}(\mathbf{k}) &\approx \mathcal{L}(\mathbf{k}_m, \mathbf{k}_{m+1}) + \mathbf{g}_m^\top (\mathbf{k} - \mathbf{k}_m) + \mathbf{g}_{m+1}^\top (\mathbf{k} - \mathbf{k}_{m+1}) + \frac{1}{2} (\mathbf{k} - \mathbf{k}_m)^\top \mathbf{H}^{11} (\mathbf{k} - \mathbf{k}_m) \\
&+ (\mathbf{k} - \mathbf{k}_m)^\top \mathbf{H}^{12} (\mathbf{k} - \mathbf{k}_{m+1}) + \frac{1}{2} (\mathbf{k} - \mathbf{k}_{m+1})^\top \mathbf{H}^{22} (\mathbf{k} - \mathbf{k}_{m+1})  
\end{aligned}  
\end{equation}  

\textbf{Taking the Derivative of $\mathcal{L}(\mathbf{k})$ with Respect to $\mathbf{k}$ and Setting to Zero}  

Taking the derivative:  
\begin{equation}  
\begin{aligned}  
\frac{\partial \mathcal{L}}{\partial \mathbf{k}} = &\mathbf{g}_m + \mathbf{g}_{m+1} + \mathbf{H}^{11} (\mathbf{k} - \mathbf{k}_m) + \mathbf{H}^{12} (\mathbf{k} - \mathbf{k}_{m+1}) + \mathbf{H}^{12\top} (\mathbf{k} - \mathbf{k}_m) + \mathbf{H}^{22} (\mathbf{k} - \mathbf{k}_{m+1})  
\end{aligned}  
\end{equation}  

Since the Hessian matrix is symmetric, i.e., $\mathbf{H}^{12} = \mathbf{H}^{21}$, we have:  
\begin{equation}  
\begin{aligned}  
\frac{\partial \mathcal{L}}{\partial \mathbf{k}} = &\mathbf{g}_m + \mathbf{g}_{m+1} + (\mathbf{H}^{11} + 2\mathbf{H}^{12} + \mathbf{H}^{22})\mathbf{k} - (\mathbf{H}^{11}\mathbf{k}_m + \mathbf{H}^{12}(\mathbf{k}_m + \mathbf{k}_{m+1}) + \mathbf{H}^{22}\mathbf{k}_{m+1})  
\end{aligned}  
\end{equation}  

Setting the derivative to zero yields the optimal condition:  
\begin{equation}  
\begin{aligned}  
(\mathbf{H}^{11} + 2\mathbf{H}^{12} + \mathbf{H}^{22})\mathbf{k}^* = &\mathbf{H}^{11}\mathbf{k}_m + \mathbf{H}^{12}(\mathbf{k}_m + \mathbf{k}_{m+1}) + \mathbf{H}^{22}\mathbf{k}_{m+1} - (\mathbf{g}_m + \mathbf{g}_{m+1})  
\end{aligned}  
\end{equation}  

\textbf{Solving for the Optimal Key Vector $\mathbf{k}^*$}  

The optimal key vector $\mathbf{k}^*$ is obtained as:  
\begin{equation}  
\begin{aligned}  
\mathbf{k}^* = &(\mathbf{H}^{11} + 2\mathbf{H}^{12} + \mathbf{H}^{22})^{-1}  (\mathbf{H}^{11}\mathbf{k}_m + \mathbf{H}^{12}(\mathbf{k}_m + \mathbf{k}_{m+1}) + \mathbf{H}^{22}\mathbf{k}_{m+1} - (\mathbf{g}_m + \mathbf{g}_{m+1}))  
\end{aligned}  
\end{equation}

\section{Comparison of AsymKV with Other New Baselines}
We also conducted additional comparisons with SnapKV~\cite{10.5555/3737916.3738638}, PyramidKV~\cite{cai2025pyramidkvdynamickvcache}, TOVA~\cite{oren-etal-2024-transformers}, D2O~\cite{wan2025textdtexto} and L\_2-Norm~\cite{devoto-etal-2024-simple}.These experiments were performed on the LongBench dataset based on the Llama3-8b-Instruct with compression context max\_length=2048.

As show in Table~\ref{tab:otherbaseline} AsymKV still demonstrates performance improvements compared to these baselines.
\begin{table*}[!htbp]
\small
\centering
\caption{Comparison of AsymKV with other new baselines on LongBench.}
\begin{adjustbox}{max width=\textwidth}
\begin{tabular}{lccccccc}
\toprule
\textbf{} & \makecell{\textbf{Single-Doc}} & \makecell{\textbf{Multi-Doc}} & \makecell{\textbf{Sum}} & \makecell{\textbf{Few-shot} } & \makecell{\textbf{Synthetic}} & \makecell{\textbf{Code}} & \textbf{Avg.} \\
\midrule
\multicolumn{8}{c}{\textbf{Llama3-8B-Instruct}} \\
\midrule
Full Context & 32.19 & 34.59 & 24.96 & 68.48 & 36.96 & 54.41 & 41.46 \\
StreamingLLM & 27.90 & 25.92 & 24.49 & 65.09 & 13.87 & 55.02 & 35.50  \\
LongCache & 28.26 & 25.64 & 24.69 & 65.75 & 15.50 & 54.65 & 35.83  \\
H$_2$O & 30.65 & 32.77 & 24.61 & 61.83 & 37.08 & 54.87 & 39.59  \\
LLMLingua-2 &26.50 & 30.80 & 24.10 & 39.30 & 22.50 & 32.20 & 29.47  \\
CaM & 30.49 & 31.48 & 24.85 & 63.83 & 37.02 & 55.46 & 39.81  \\
TOVA & 31.82 & 27.94 & 24.57 & 64.34 & 19.29 & 54.06 & 37.04 \\
L2 & 30.18 & 27.41 & 24.70 & 63.29 & 37.34 & 51.78 & 38.43 \\
D2O & 30.81  & 32.87  & 24.64  & 67.42 & 36.67 & 	56.49 & 40.85  \\
SnapKV & 32.17 & \textbf{34.20} & 25.28 & \textbf{68.57} & 37.21 & 53.30 & 41.36 \\
PyramidKV & 31.79 & 34.02 & 25.44 & 68.57 & 37.24 & 54.97 & 41.49 \\
\cellcolor{highlight}{\textbf{AsymKV}} & \cellcolor{highlight}{\textbf{34.45}} & \cellcolor{highlight}{33.64} & \cellcolor{highlight}{\textbf{26.17}} & \cellcolor{highlight}{67.94} & \cellcolor{highlight}{\textbf{38.66}} & \cellcolor{highlight}{\textbf{56.61}} & \cellcolor{highlight}{\textbf{42.32}} \\
\bottomrule
\end{tabular}
\end{adjustbox}
\label{tab:otherbaseline}
\end{table*}

\section{Licenses for Existing Assets}  
We list the assets used in this paper and their licenses below:
\begin{itemize}
    \item ~\cite{touvron2023llama},llama2
    \item ~\cite{dubey2024llama},llama3
    \item ~\cite{jiang2023mistral7b},Apache 2.0 License
    \item ~\cite{yang2024qwen2technicalreport},Apache 2.0 License
    \item ~\cite{bai2024longbench},MIT License
    \item ~\cite{li2023long},Apache 2.0 License
    \item ~\cite{yang2018hotpotqa},CC BY-SA 4.0
    \item ~\cite{bai2024longbench2},MIT License
    \item ~\cite{an-etal-2024-l},GNU General Public License v3.0
    \item ~\cite{vicuna2023},llama2
    \item ~\cite{xiao2023streamingllm},MIT License
\end{itemize}